\documentclass{article} 

\PassOptionsToPackage{table}{xcolor}

\usepackage{iclr2027_conference,times}

\usepackage[utf8]{inputenc}
\usepackage[T1]{fontenc}
\usepackage{hyperref}
\usepackage{url}
\usepackage{xcolor}
\usepackage{microtype}
\usepackage{nicefrac}

\usepackage{etoc}
\usepackage{titletoc}

\usepackage{amsmath}
\usepackage{amssymb}
\usepackage{amsfonts}
\usepackage{mathtools}
\usepackage{amsthm}

\usepackage{graphicx}
\usepackage{subcaption}
\usepackage{booktabs}
\usepackage{threeparttable}
\usepackage{tablefootnote}
\usepackage{multirow}
\usepackage{float}
\usepackage{wrapfig}
\usepackage{colortbl}
\usepackage{caption}

\usepackage{algorithm}
\usepackage{algorithmic}

\usepackage{listings}
\usepackage{inconsolata}

\usepackage[most]{tcolorbox}
\tcbuselibrary{listings,skins,breakable}

\usepackage{colortbl}
\usepackage[table]{xcolor}
\usepackage[capitalize,noabbrev]{cleveref}

\usepackage{tikz}
\usetikzlibrary{positioning,calc,fit,arrows.meta,shapes.misc}

\usepackage{pifont}

\usepackage[disable,textsize=tiny]{todonotes}

\newcommand{\mstd}[2]{#1$\pm$\text{\scriptsize #2}}
\newcommand{\bmstd}[2]{\textbf{#1}$\pm$\text{\scriptsize #2}}

\newcolumntype{G}{!{\color{black!45}\vrule width 0.35pt}}
\theoremstyle{plain}

\theoremstyle{definition}

\theoremstyle{remark}

\newcommand{\thickhline}{\noalign{\hrule height 0.8pt}}

\newcolumntype{v}{!{\vrule width 0.25pt}}
\title{Directing to Diversify: Multi-Directional\\ Search for Automatic Heuristic Design}

\iclrfinalcopy

\author{
  Jianghan Zhu$^{1}$, 
  Jianan Zhou$^{2}$
  , Fei Liu$^{3,4}$, 
  Ni Zhang$^{1}$,\\
  \textbf{ Yaoxin Wu$^{5}$,
  Zhuoyi Lin$^{6}$, 
  Zhiguang Cao$^{1}$} \\ 
  \noalign{\vskip 3pt}
  $^1$Singapore Management University \quad $^2$Nanyang Technological University \\
  $^3$University of Zurich \quad $^4$ETH Zurich \quad $^5$Eindhoven University of Technology \quad $^6$A*STAR \\
}

\begin{document}

\maketitle

\begin{abstract}
    Large language models (LLMs) have demonstrated strong potential to automate heuristic design for NP-hard combinatorial optimization (CO) tasks without human intervention. Recent methods predominantly rely on evolutionary computation (EC) frameworks, where heuristic diversification is mainly induced through EC-like operators or downstream selection, while the direction of heuristic transformation remains largely unharnessed. In this paper, we propose a multi-directional \textbf{P}erturbation--\textbf{R}ewrite \textbf{S}earch (\textbf{PRS}) framework for automatic heuristic design that factorizes heuristic transformation through directional search primitives at the generation level. Specifically, PRS operates on heuristic ideas through three complementary directions: expansion toward richer and more expressive designs, compression toward simpler and less specialized designs, and steering for fine-grained refinement, together enabling broader and more sustained exploration of diverse regions in the heuristic design space. PRS further organizes candidate exploration within a breadth-first search-based process with adaptive operator control, which ensures thorough candidate utilization and sustained exploration, and maintains strong search dynamics while preventing premature convergence. Extensive empirical results across diverse CO tasks show that PRS consistently discovers high-quality heuristics, demonstrating the effectiveness and generality of explicitly directional exploration design.
\end{abstract}

\section{Introduction}
\label{sec:intro}

Automatic heuristic design (AHD) aims to automatically discover high-performing heuristic algorithms for a wide range of optimization problems~\citep{he2011heuristic,dahmani2024solving,novikov2025alphaevolve}. Despite its importance, traditional AHD methods, such as genetic programming, remain fundamentally constrained by their reliance on manually designed, problem-specific operators, which limits scalability and restricts the exploration of the heuristic search space. The recent emergence of large language models (LLMs) has opened a promising new direction for AHD~\citep{liu2024systematic}, enabling automated heuristic generation through LLM-based code synthesis and reasoning.

Most existing LLM-based AHD approaches build upon optimization frameworks (e.g., evolutionary computation (EC)), where heuristics are iteratively improved using LLMs~\citep{romera2024mathematical,liu2024evolution,ye2024reevo,zheng2025monte,shi2026generalizable}. Maintaining diverse heuristic candidates is important for preventing premature concentration around a narrow family of designs and for preserving access to alternative regions that may become useful at different stages of search. Recent works have attempted to enhance the search process with more structured control, such as diversity-aware population management~\citep{dat2025hsevo} and feature-assisted partitioning of the abstract heuristic search space~\citep{hu2025partition}. For example, EvoX~\citep{liu2026evox} adaptively evolves the search strategy that governs parent selection and candidate variation, dynamically switching among refinement, structural variation, and free-form variation to balance exploitation and diversification according to search progress.


Despite these advances, existing approaches mainly encourage diversification through population management, candidate selection, similarity measures, or search-space partitioning. Such mechanisms regulate which candidates are retained, preferred, or grouped after or around generation, but they do not explicitly structure the generation process itself around different directions of heuristic change. As a result, heuristic diversification is typically an emergent consequence of generic mutation, crossover, reflection, or novelty pressure, rather than a directly controlled search primitive. This leaves an important gap: existing AHD frameworks lack explicit generation primitives for deliberately diversifying heuristic ideas along complementary semantic directions.


In this work, we argue that diversification should be controlled directly at the level of heuristic generation. To this end, we propose \textbf{Perturbation--Rewrite Search (PRS)}, a multi-directional search framework that explicitly factorizes heuristic generation into complementary search primitives over the idea space. PRS introduces three perturbation strategies, i.e., \emph{expansion}, \emph{compression}, and \emph{steering}, which deliberately transform a parent heuristic toward richer, simpler, and locally refined designs, respectively. Rather than using diversity only as a population property or selection signal, PRS controls \emph{how} new heuristic ideas are produced before they are realized as executable implementations. These abstraction-level perturbations are then instantiated through an LLM-based rewriting process that translates the modified heuristic specification into executable code while preserving its intended decision logic. Together, this primitive-level directional diversification enables broader and more sustained coverage of structural regions in the heuristic space.


To further enhance search effectiveness, PRS organizes candidate exploration within a breadth-first search (BFS) framework. By expanding multiple candidates layer by layer, PRS improves candidate utilization while preventing early domination by a small subset of heuristics. This search strategy is coupled with an adaptive exploration control mechanism, enabling sustained exploration and robustness against stagnation.

    
    

Our contributions are summarized as follows:
\begin{itemize}
    \item We introduce Perturbation--Rewrite Search (PRS) for automatic heuristic design, which makes directional heuristic transformation an explicit generation-level search primitive for structured exploration of the heuristic idea space.
    
    \item We develop a multi-directional perturbation mechanism with expansion, compression, and steering, which deliberately transforms heuristic ideas toward richer, simpler, and locally refined designs, respectively. PRS further integrates bounded BFS-based search and adaptive operator control to allocate search effort across these complementary directions.
    
    \item We comprehensively evaluate PRS across 11 combinatorial optimization problem settings and show that it consistently discovers high-quality heuristics, demonstrating strong effectiveness and generality compared to state-of-the-art LLM-based AHD methods.
\end{itemize}

\section{Preliminary}
\label{preliminary}
\textbf{AHD.}
Let $\mathcal{P}$ denote a problem class, and let $\mathcal{S}(p)$ be the feasible solution set for an instance $p\in\mathcal{P}$.
A heuristic space is defined as $\mathcal{H}$, where each heuristic $h\in\mathcal{H}$ induces a mapping
$h:\mathcal{P}\rightarrow \bigcup_{p\in\mathcal{P}}\mathcal{S}(p)$, and for any $p\in\mathcal{P}$, $h(p)\in \mathcal{S}(p)$. 
For a given validation set $\mathcal{D}\subseteq\mathcal{P}$ and an objective function $f:\mathcal{S}\rightarrow\mathbb{R}$,
the goal of AHD is to find an optimal heuristic $h^{\star}\in\mathcal{H}$ that maximizes expected performance $g$:
\begin{equation}
h^{\star} = \arg\max_{h\in\mathcal{H}} g(h) = \arg\max_{h\in\mathcal{H}} \mathbb{E}_{x \sim \mathcal{D}}f(h(x)).
\label{eq:ahd_objective}
\end{equation}

\textbf{LLM-based AHD.} LLM-based AHD leverages LLMs to navigate the typically vast heuristic space $\mathcal{H}$ in CO. In practice, the designed heuristic often constitutes only a key component of a broader optimization algorithm.
For example, in Travelling Salesman Problem (TSP), existing LLM-based AHD methods \citep{liu2024evolution} typically focus on designing step-by-step constructive heuristics that select the next node based on the current partial tour. Beyond constructive frameworks, recent studies have also explored learning heuristics that update weighted distance matrices within the broader Guided Local Search (GLS) algorithm, thereby more effectively guiding local search decisions \citep{ye2024reevo,zheng2025monte}.

A prominent line of LLM-based AHD methods follows an evolutionary paradigm \citep{liu2023algorithm,romera2024mathematical,liu2024evolution,ye2024reevo}. Using EC-style operators such as crossover and mutation, these methods progressively explore the heuristic space, with more recent extensions further strengthening this paradigm. 
For example, HiFo-Prompt \citep{chen2025hifo} introduces hierarchical prompt-level guidance through hindsight--foresight control and knowledge distillation for faster and more efficient heuristic search.

Despite differences in evolution strategies, these frameworks remain rooted in EC-like search, where search operators typically recombine multiple heuristics in crossover to inherit parental features, together with mutation-style operators to promote diversification. Consequently, the direction of heuristic change is largely uncontrolled by the search mechanism itself: different transformation patterns may emerge during evolution, but they are not explicitly harnessed as distinct search primitives. In contrast, PRS directly structures heuristic generation through complementary transformation directions, enabling multi-directional exploration without relying primarily on recombination of existing structures.



\textbf{Breadth First Search.}
Breadth-first search is a fundamental graph traversal strategy that explores states layer by layer by maintaining a frontier of candidate states at each depth. 
Recent work such as Tree-of-Thoughts \citep{yao2023tree} instantiates BFS in the space of intermediate reasoning trajectories, where each layer corresponds to a single reasoning step and the frontier is pruned to a fixed breadth. In a generic BFS procedure, the search starts from an initial state and expands all states in the current frontier using a predefined expansion operator. The resulting states are then scored, and a subset of top-ranked states is retained as the frontier for the next layer. This process repeats until a stopping criterion, such as a maximum search depth or a computational budget, is reached.

\section{Methodology}
\label{sec:methodology}

In this section, we introduce PRS, a multi-directional perturbation--rewrite search framework that makes directional heuristic transformation an explicit search primitive for automatic heuristic design.
PRS explores the heuristic space by applying multi-directional perturbations to heuristic ideas and then rewriting the corresponding implementations to realize the modified heuristic as executable candidates. 
At the search level, PRS maintains a heuristic pool and organizes candidate expansion in a breadth-first manner with selective pruning to enhance candidate utilization.
In addition, PRS incorporates an adaptive exploration controller to regulate the exploration--exploitation trade-off and mitigate premature convergence during heuristic search.

\begin{wrapfigure}{r}{0.5\columnwidth}
  \vspace{-2mm}
  \centering
  \includegraphics[width=0.5\columnwidth]{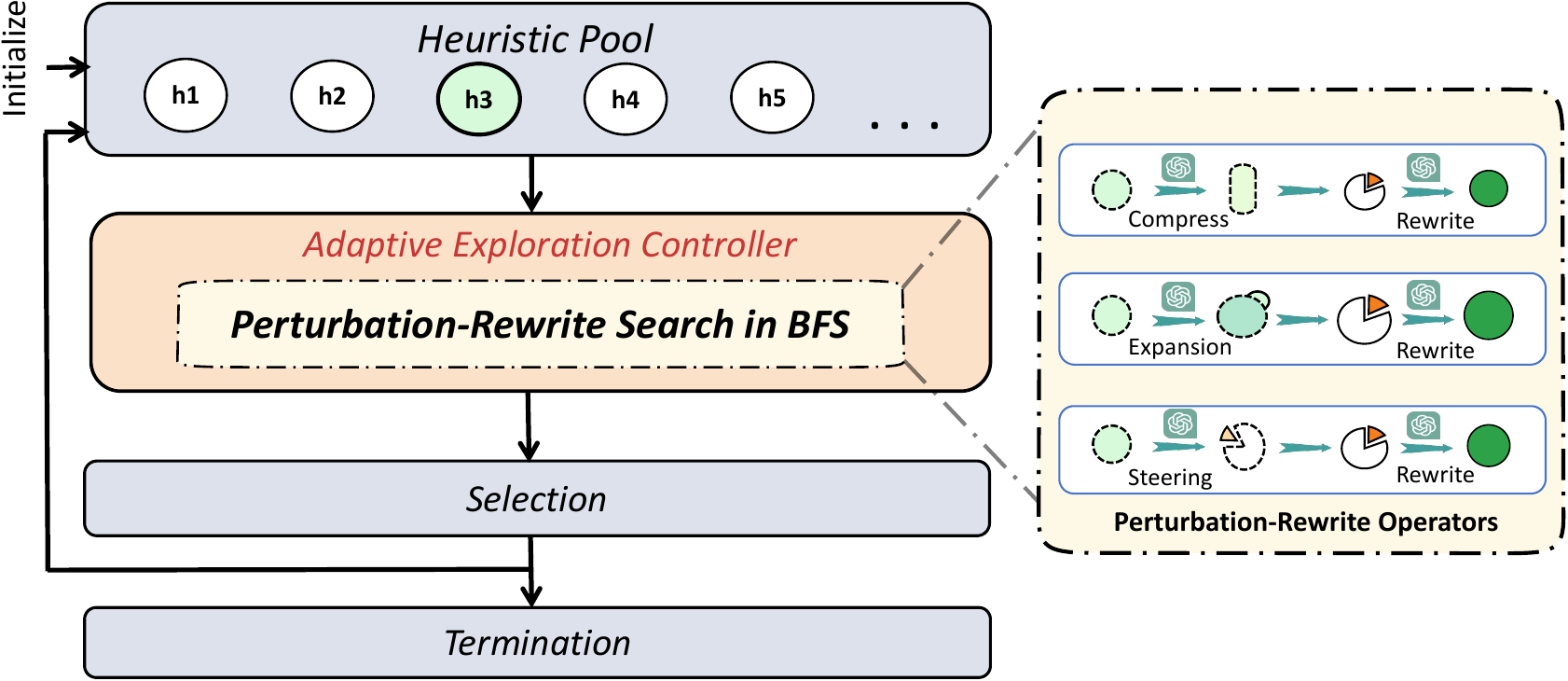}
  \caption{An illustration of PRS pipeline.}
  \label{fig:PRS_framework}
  \vspace{-3mm}
\end{wrapfigure}
\begin{figure*}[t]
\vspace{1mm}
  \centering
  \includegraphics[width=0.9\textwidth]{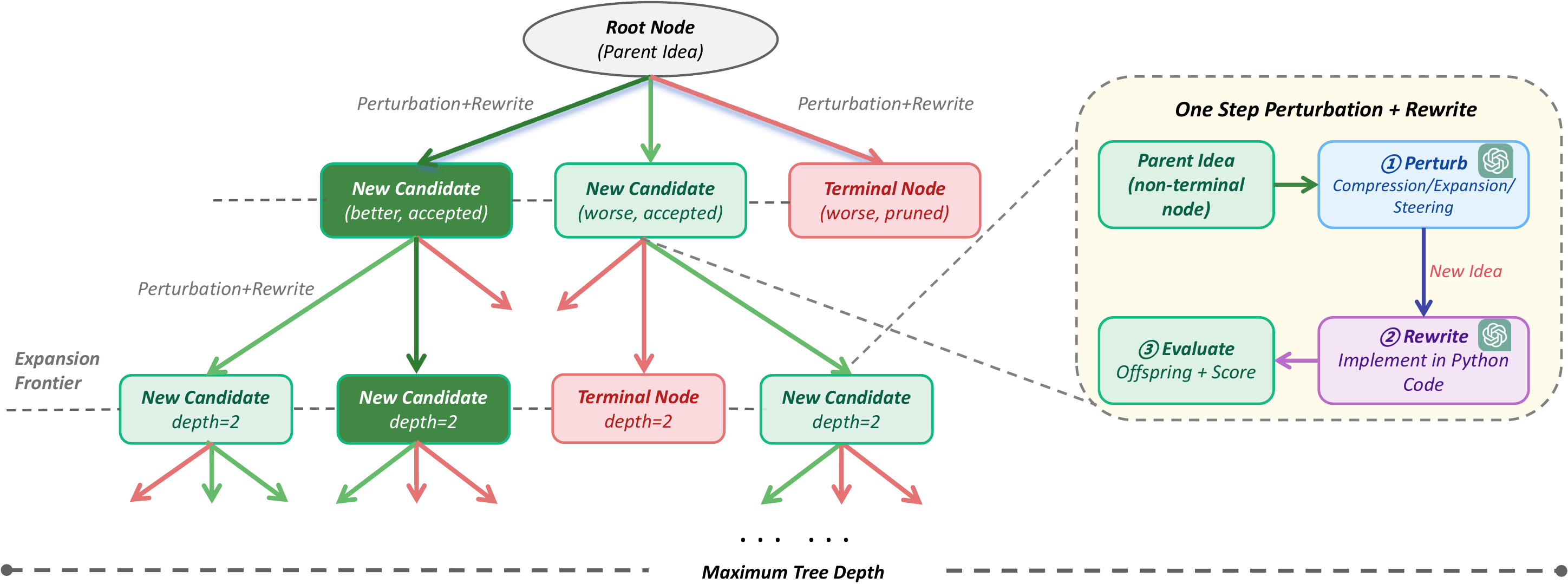}
  \caption{An overview of BFS-based Heuristic Search.
  Starting from a root parent heuristic with its corresponding idea, PRS performs a BFS-style expansion in which new offspring ideas are generated via multi-directional perturbation and then realized as executable code through rewriting. The offspring codes are then evaluated and selectively retained in a fixed-size frontier at each depth. 
  Terminal nodes correspond to pruned or non-extendable offspring.
  \textcolor{green!40!black}{Dark green} boxes denote heuristics that outperform the root parent, while \textcolor{green!70!black}{light green} (or \textcolor{red!60!white}{light red}) boxes represent worse-than-root offspring that are accepted (or rejected) for inclusion in the next frontier. Each parent heuristic across BFS paths retains opportunity for further expansion, preventing early domination by a single trajectory.
  The beam size is set to $\kappa=3$ for illustration.}
  \label{fig:PRS_BFS}
  \vspace{-1mm}
\end{figure*}

\textbf{PRS Pipeline.} 
The overall pipeline of PRS is illustrated in Fig.~\ref{fig:PRS_framework}. 
Specifically, PRS starts the heuristic search by initializing a heuristic pool, which maintains a set of parent heuristics with a selection mechanism, and terminates after a certain condition. 
We introduce a novel multi-directional perturbation--rewrite operator that explores the heuristic space through complementary transformation directions. It includes expansion, compression, and steering, which transform heuristic ideas along complementary directions in the heuristic design space. These operators are used to expand parent nodes within a BFS framework, as illustrated in Fig \ref{fig:PRS_BFS}. Regulated by an adaptive exploration controller, these operators jointly facilitate an effective search for the optimal heuristic $h^*$ for a given task. 
After each completed BFS tree, random sampling is applied to the heuristic pool to promote diversity and retain temporarily inferior yet potentially promising candidates, while always preserving the current best heuristic.
The complete PRS algorithm and associated prompts are presented in Algorithm \ref{alg:PRS} and Appendix \ref{sec:app_propmt}, respectively. In the following subsections, we describe the perturbation and rewrite process, the BFS-based search structure, and the adaptive exploration controller in detail.


\subsection{Perturbation-Rewrite Operator}
\label{sec:Perturbation-rewrite_operators}
PRS introduces a multi-directional perturbation--rewrite process composed of a perturbation step followed by an LLM-based rewrite step. Rather than directly modifying executable heuristics through carefully crafted EC operators, PRS applies directional perturbations to heuristic ideas and subsequently rewrites the corresponding implementations to realize the modified heuristics as executable candidates. This design separates the intended algorithmic modification from its low-level implementation details, enabling controlled changes in heuristic structure while preserving the executability and semantic coherence required for black-box heuristic evaluation. We denote $h_p$ as the parent heuristic used for perturbation--rewrite search, and elaborate the perturbation and rewriting strategies below.

\textbf{Multi-directional Perturbation.} The goal of multi-directional perturbation is to explore the heuristic space through complementary transformation directions, thereby overcoming the localized perturbation nature of traditional mutation operators.
This design is motivated by the fact that heuristic quality does not vary monotonically with structural complexity: overly simple heuristics may lack expressive power, whereas overly complex ones may become bulky, computationally expensive, or over-fit to in-distributional data.
Accordingly, PRS employs three complementary perturbation actions that address different search needs: expansion increases heuristic expressiveness, compression counteracts excessive specialization, and steering provides refinement around promising heuristic structures.

\emph{Expansion $(\Phi^{\text{expn}}_{\theta})$.} The expansion operator explicitly prompts LLMs to enrich the parent heuristic by introducing additional decision signals, interactions, conditional rules, or adaptive mechanisms, deliberately increasing its structural complexity and expressive capacity. Specifically, expansion instructs LLMs to inject new structural components, such as additional scoring terms or conditional logic. The operator further specifies how these additions affect heuristic decisions rather than simply appending independent components. From an optimization perspective, expansion biases heuristic structures toward more elaborate and expressive programmatic forms, such as deeper control flow or richer operator compositions, thereby increasing its expressive capacity and conceptual richness.

\emph{Compression $(\Phi^{\text{comp}}_{\theta})$.} Complementary to expansion, PRS employs a compression operator that biases LLMs toward reducing the structural complexity of the current parent heuristic by removing redundant mechanisms or over-specialized components that may hinder out-of-distribution (OOD) generalization (e.g., structures tightly coupled to specific problem sizes or data distributions).
Intuitively, compression may disrupt the established advantages of the parent heuristic, as simplification can often lead to short-term performance degradation. However, from a broader perspective, it may mitigate the accumulation of dominant structural priors and excessive complexity, acting as a potential refactoring mechanism in the heuristic space that counteracts over-specialization and promotes more robust heuristic representations.

\emph{Steering $(\Phi^{\text{steer}}_{\theta})$.} 
While expansion and compression enable substantial shifts toward richer or simpler heuristic designs, they are less suited for fine-grained exploitation of an already promising heuristic idea. Steering complements these directional perturbations by performing localized refinements while approximately preserving the overall heuristic structure, allowing PRS to further exploit promising regions without inducing another substantial structural drift.

Together, expansion, compression, and steering constitute a multi-directional perturbation mechanism that enables PRS to explore both richer and more generalized heuristics across complementary regimes.
The perturbation process can be formally expressed as
\begin{equation}
\tilde{z} \;=\; \Phi^{(o)}_{\theta}\!\left(z_p \mid \mathcal{C}\right),
\end{equation}
where $z_p$ denotes the heuristic idea of the parent heuristic, $\tilde{z}$ denotes the perturbed heuristic idea, $\mathcal{C}$ contains the task and function descriptions, and
$\Phi_{\theta}^{(o)} \in \{\Phi^{\text{steer}}_{\theta}, \Phi^{\text{expn}}_{\theta}, \Phi^{\text{comp}}_{\theta}\}$ denotes the selected perturbation operator.
At each perturbation step, one of these actions is adaptively selected according to its recent search utility.

\textbf{Rewrite.}
Following multi-directional perturbation, the perturbed offspring idea specifies the intended algorithmic change but does not directly constitute an executable heuristic.
The rewrite step therefore prompts the LLM to realize the offspring idea by rewriting the complete parent implementation, preserving existing mechanisms when consistent with the target idea. PRS independently rewrites the same offspring idea into multiple offspring candidates according to the beam size $\kappa$, allowing different executable realizations of the same heuristic perturbation.
The rewrite process can be formally expressed as:
\begin{equation}
h'_j \;\sim\; \Psi_{\theta}\!\left(h_p,\,\tilde{z},\,o \mid \mathcal{C}\right),
\qquad j=1,\ldots,\kappa,
\end{equation}
where $\Psi_{\theta}$ denotes the LLM-based rewrite operator, $h_p$ is the parent heuristic implementation, $\tilde{z}$ is the perturbed heuristic idea produced by operator $o$, and $\kappa$ denotes the beam size.

\subsection{BFS-based Heuristic Search}

We employ a bounded BFS-style search mechanism in conjunction with the perturbation--rewrite operator, as BFS naturally promotes candidate utilization through layer-by-layer expansion and prevents early domination by a small subset of heuristics.
At the root level, one parent heuristic is selected from the current heuristic pool using an $\epsilon$-greedy strategy based on its performance.
Let $h^*$ denote the best heuristic in the current pool $\mathcal{P}$. At the root level of each BFS tree, $h^*$ is selected as the root parent with probability $1-\epsilon$, while each of the remaining heuristics is selected with probability $\frac{\epsilon}{|\mathcal{P}|-1}$. We set $\epsilon=0.5$ in all experiments to balance exploitation of the current best heuristic and exploration of alternative parents.

As shown in Fig.~\ref{fig:PRS_BFS}, for each non-terminal parent at the current BFS frontier,  one perturbation operator is selected to generate offspring candidates through the perturbation--rewrite process.
After all parents in the current frontier have been expanded, the resulting offspring at the same depth are evaluated based on their average performance on the validation dataset.
A subset of candidates is then retained as the next frontier through softmax-based sampling conditioned on their performance scores.
For a candidate set $\tilde{\mathcal{F}}$ at the current BFS depth, the probability of retaining a candidate $h_i$ is defined as:
\begin{equation}
p(h_i) =
\begin{cases}
1, 
& \text{if } h_i = \displaystyle\arg\max_{h \in \tilde{\mathcal{F}}} g(h), \\[6pt]
\operatorname{Softmax}(g(h_i)),
& \text{otherwise}.
\end{cases}
\label{eq:softmax_frontier}
\end{equation}



Each retained offspring serves as a parent for the next BFS layer.
The BFS expansion terminates when the maximum tree depth is reached or the remaining evaluation budget is exhausted.
After each completed BFS tree, the surviving frontier candidates are immediately merged with the current heuristic pool, where the best heuristic is preserved and the remaining pool members are sampled randomly to maintain diversity.

\subsection{Adaptive Exploration Controller}
\label{sec:adaptive_exploration_controller}
The BFS-based heuristic search with the perturbation--rewrite operator constitutes the core of PRS. In this subsection, we introduce an adaptive exploration controller that serves as an auxiliary mechanism to balance exploration and exploitation, and to mitigate search stagnation in local optima.

\textbf{Adaptive Operator Control.}
In evolutionary search, the balance between exploration and exploitation critically influences both search breadth and convergence quality. Excessive exploration can lead to random drift and insufficient refinement of promising elites, whereas overly conservative updates or exploitation may cause the search to collapse into a narrow region of the solution space. Previous studies have shown that static control policies are generally suboptimal in different search phases \citep{vcrepinvsek2013exploration}.

To address this issue, PRS adaptively selects among expansion, compression, and steering using a UCB-based criterion according to their recent search utility. For an operator $o\in\mathcal{O}$, its selection score at evaluation step $t$ is defined as
\begin{equation}
U_t(o)
=
\bar{R}_t(o)
+
c_t
\sqrt{\frac{\log N_t}{N_t(o)}},
\label{eq:operator_ucb}
\end{equation}
where $\bar{R}_t(o)$ denotes the mean reward of operator $o$ within a recent sliding window, $N_t(o)$ is the number of times $o$ has been selected within the window, and $N_t$ is the total number of recorded operator selections. For each parent expansion, the operator reward is computed as
$R_t(o)=\max\left\{0,\frac{g(h'_*)-g(h_p)}{|g(h_p)|}\right\}$,
where $h'_*$ is the best offspring generated from the corresponding parent.
Operators that have not yet been observed in the current window are prioritized before applying the UCB criterion. The exploration coefficient $c_t$ controls the trade-off between operator exploration and reward-guided selection, and follows a linear annealing schedule as a function of the evaluation step $t$. Let $T$ denote the maximum evaluation step. We define:
\begin{equation}
c_t = c_0 + (c_T - c_0)\frac{t}{T},
\label{eq:ucb_anneal}
\end{equation}
where $c_0$ and $c_T$ are the initial and final exploration coefficients, respectively. This schedule encourages broader exploration of different perturbation directions at the beginning and gradually shifts toward exploiting operators with higher observed utility.

\textbf{Stagnation-Aware Restart.}
While adaptive operator control improves the overall trade-off between exploration and exploitation, it cannot guarantee escape from stagnation in local optima \citep{dat2025hsevo}.
As the search progresses, repeatedly exploiting successful perturbation directions and promising parents may gradually concentrate the search around a narrow set of heuristic structures, making further improvements increasingly difficult.
This behavior arises because the best-performing heuristics can repeatedly attract search effort within their local neighborhoods, leading to the proliferation of structurally similar candidates.
As a result, the search may enter a \emph{plateau} regime, where heuristic diversity diminishes due to premature convergence or inbreeding \citep{eshelman1991preventing}. 

To address this issue, PRS triggers a stagnation-aware restart when no improvement in the global best heuristic is observed for a predefined number $\mathcal{C}$ of heuristic evaluations.
Once activated, two coordinated actions are performed:
a) \emph{Elite Restart}, where the heuristic pool is reset to retain only the globally best heuristic discovered so far; and
b) \emph{Operator Reheating}, where the recent operator reward window is reset and the UCB exploration coefficient is temporarily restored to its initial value $c_0$, shifting operator selection back toward exploration before gradually returning to reward-guided exploitation. Together, these mechanisms enable the heuristic search to escape the current \emph{plateau} by reintroducing diversity into the heuristic pool while ensuring smooth convergence through restarting from the elite heuristic. The complete PRS algorithm and associated prompts are presented in Algorithm \ref{alg:PRS} and Appendix \ref{sec:app_propmt}, respectively.

\section{Experiments}
\label{sec:experiments}

This section empirically evaluates PRS on some typical NP-hard COPs. The definitions of these COPs are listed in the Appendix \ref{sec:definition of tasks}. Importantly, these problems are also applied to a wide range of general frameworks, including step-by-step constructive, guided-local search (GLS) and ant colony optimization (ACO). Details of these general frameworks are included in the Appendix \ref{sec: evaluated framework}. The metrics used to measure the performance of each method primarily include the average objective and the average gap to the reference value (e.g., best-known solution) of each problem size on the test set. Our evaluated problem setting covers a wide range of problem types and general frameworks: 1) ACO:  Capacitated Vehicle Routing Problem (CVRP), Orienteering Problem (OP) (both black-box and white-box), Multiple Knapsack Problem (MKP) (black-box); 2) Step-by-step Construction: Knapsack Problem (KP), Online Bin Packing Problem (OBPP), TSP, Admissible Set Problem (ASP); 3) GLS: TSP, Flow Shop Scheduling Problem (FSSP).


\textbf{Implementation Setups.}
To maximize the effectiveness of BFS, we adopt a beam search with beam size $\kappa=3$ and maximum tree depth $d=3$, and set the frontier size (i.e., the maximum number of parent nodes for the next layer) to $fs=3$. For both validation datasets and testing datasets, we follow the procedures and configurations of the MCTS-AHD \citep{zheng2025monte} and use their random seeds to generate all datasets. In addition, to ensure a fair comparison, we adopt the same seed heuristics as MCTS-AHD. For all problems except OBPP, we set the number of evaluated heuristics to $T=1000$ and constrain all LLM-based AHD methods to a maximum budget of 5,000 LLM queries.
The size of the heuristic pool is set to 10 for all problems. Comprehensive hyperparameter settings and experimentation setups can be found in the Appendix \ref{sec:evaluation_details}. All experiments were conducted on a machine equipped with a single Intel Core i9-12900KF CPU. 

\textbf{Baselines.} We compare PRS with the following baselines. They are mainly composed of four categories: (1) hand-crafted heuristics such as Greedy-Construct (GC) \citep{rosenkrantz1977analysis}, Best-Fit for OBPP \citep{kenyon1996best}, GPUTA \citep{gupta1971functional}, ACO \citep{blum2005ant,dorigo2006ant} algorithm for CVRP, OP, MKP, etc. (2) Traditional heuristic solvers, including ILS1 \citep{stutzle1998applying}, OR-Tools, etc. (3) Neural Combinatorial Optimization (NCO) methods including POMO \citep{kwon2020pomo}, DeepACO \citep{ye2023deepaco}, etc.
(4) Existing LLM-based AHD methods. They are: FunSearch \citep{romera2024mathematical}, EoH \citep{liu2024evolution}, ReEvo \citep{ye2024reevo}, HSEvo \citep{dat2025hsevo} and MCTS-AHD \citep{zheng2025monte}. All methods are run with the same initial seed heuristics and tested on identical validation datasets $D$. Unless otherwise stated, all LLM-based AHD methods are evaluated using GPT-5-nano and mean performance is based on three independent runs.

\subsection{Comparison on ACO Framework} 

The ACO framework is a classical metaheuristic mechanism for solving combinatorial optimization problems by simulating pheromone-guided stochastic construction \citep{dorigo2006ant,kim2024ant}. By updating the pheromone matrix $\tau$ to determine the edge priority and crafting a more complicated heuristic matrix $\eta$ to guide ants’ transition probabilities and assess the constructed tours, ACO can iteratively drive the solution towards optimal or near-optimal regions. As shown in Table \ref{tab:performance_cvrp_op}, under GPT-5-nano, PRS consistently outperforms all compared LLM-based AHD baselines on the out-of-distribution settings of both CVRP and OP. On the in-distribution settings, PRS also achieves the best performance on CVRP and remains competitive on OP, outperforming EoH and MCTS-AHD. These results suggest that PRS is particularly effective in discovering heuristics with stronger out-of-distribution generalization under the ACO framework.
\begin{table*}[h]
\centering
\caption{
Performance of ACO-based heuristics on CVRP and OP. All
methods are evaluated on three test sets of 64 randomly generated instances each, following the setup
in~\cite{dat2025hsevo,zheng2025monte}. Test sets whose scales match the validation sets are underlined. Optimal solutions are approximated using DeepACO with significantly
more ants and iterations than those in the baselines. The best-performing method is shaded, and each test set’s overall best result is in bold. 
}
\label{tab:performance_cvrp_op}
\normalsize
\setlength{\tabcolsep}{3.5pt}
\renewcommand{\arraystretch}{1.0}

\resizebox{\textwidth}{!}{%
\begin{tabular}{l|cc|cc|cc|cc|cc|cc}
\noalign{\hrule height 1.2pt} 
\multicolumn{1}{c|}{Task}
& \multicolumn{6}{c|}{\textbf{CVRP}}
& \multicolumn{6}{c}{\textbf{OP}} \\
\hline
\multicolumn{1}{c|}{N=}
& \multicolumn{2}{c|}{\underline{N=50}}
& \multicolumn{2}{c|}{N=100}
& \multicolumn{2}{c|}{N=200}
& \multicolumn{2}{c|}{\underline{N=50}}
& \multicolumn{2}{c|}{N=100}
& \multicolumn{2}{c}{N=200} \\
\hline
\multicolumn{1}{c|}{Method}
& Obj$\downarrow$ & Gap$\downarrow$
& Obj$\downarrow$ & Gap$\downarrow$
& Obj$\downarrow$ & Gap$\downarrow$
& Obj$\uparrow$ & Gap$\downarrow$
& Obj$\uparrow$ & Gap$\downarrow$
& Obj$\uparrow$ & Gap$\downarrow$ \\
\hline

DeepACO & 8.888 & -- & 14.932 & -- & 27.159 & -- & 19.867 & -- & 36.392 & -- & 63.380 & -- \\
ACO     & 18.282 & 105.69\% & 30.030 & 101.11\% & 48.582 & 78.88\%
        & 13.570 & 31.70\%  & 24.149 & 33.64\%  & 37.109 & 41.45\% \\
\hline
\multicolumn{13}{c}{LLM-based AHD (GPT-5-nano)} \\
\hline
EOH     & 9.320 & 4.85\% & 16.511 & 10.57\% & 29.603 & 9.00\%
        & 15.162 & 23.68\% & 29.699 & 18.39\% & 53.497 & 15.59\% \\
ReEvo
& 9.186 & 3.35\% & 16.020 & 7.29\% & 28.782 & 5.98\%
& \cellcolor{gray!20}\textbf{15.290} & \cellcolor{gray!20}\textbf{23.04\%}
& 30.820 & 15.31\% & 54.459 & 14.08\% \\
MCTS-AHD
       & 9.025 & 1.54\% & 15.974 & 6.98\% & 28.572 & 5.20\%
       & 15.192 & 23.53\% & 30.752 & 15.50\% & 54.718 & 13.67\% \\
Ours
& \cellcolor{gray!20}\textbf{8.958} & \cellcolor{gray!20}\textbf{0.79\%}
& \cellcolor{gray!20}\textbf{15.576} & \cellcolor{gray!20}\textbf{4.31\%}
& \cellcolor{gray!20}\textbf{28.228} & \cellcolor{gray!20}\textbf{3.93\%}
& 15.247 & 23.25\%
& \cellcolor{gray!20}\textbf{30.924} & \cellcolor{gray!20}\textbf{15.03\%}
& \cellcolor{gray!20}\textbf{55.821} & \cellcolor{gray!20}\textbf{11.93\%} \\
\noalign{\hrule height 1.2pt} 
\end{tabular}%
}
\end{table*}

\subsection{Comparison on KP and OBPP} 
The step-by-step constructive framework incrementally builds a feasible solution by repeatedly selecting and adding one component at a time, where each decision is guided by a problem-specific heuristic and must satisfy the corresponding feasibility constraints \citep{asani2023computation}. For KP, the heuristic scores the remaining items based on their values, weights, and current residual capacity, and iteratively selects the next item until no feasible additions are possible. 
\begin{table}[t]
\vspace{1.5mm}
\small
\centering
\caption{
\textbf{Online bin packing problem results}.
The first four problem scales are used during validation, and the remaining are used for held-out test. Notations of problem scales are abbreviated, e.g., 1k\_500 indicates 1,000 items and a fixed bin capacity 500.
}
\label{tab:obpp}

\setlength{\tabcolsep}{4pt}
\renewcommand{\arraystretch}{1.15}

\begin{tabular}{l|ccccccc}
\noalign{\hrule height 1.2pt}
\multicolumn{1}{c|}{Test Sets}
& \underline{1k\_100}
& \underline{1k\_500}
& \underline{5k\_100}
& \underline{5k\_500}
& 10k\_100
& 10k\_500
& AVG \\
\hline
\multicolumn{8}{c}{LLM-based AHD (GPT-5-nano)} \\
\hline
Best Fit
& 4.77\% & 0.25\% & 4.31\% & 0.55\% & 4.05\% & 0.47\% & 2.40\% \\
First Fit
& 5.02\% & 0.25\% & 4.65\% & 0.55\% & 4.36\% & 0.50\% & 2.56\% \\
\hline
FunSearch
& 2.45\% & 0.66\% & 1.30\% & \textbf{0.25\%} & 1.05\% & \textbf{0.21\%} & 0.99\% \\
EoH
& 2.69\% & 0.25\% & 1.63\% & 0.53\% & 1.47\% & 0.45\% & 1.17\% \\
ReEvo
& 3.13\% & 0.25\% & 1.78\% & 0.55\% & 2.63\% & 0.50\% & 1.47\% \\
HSEvo
& 2.24\% & 0.50\% & 1.46\% & 0.38\% & 1.30\% & 0.36\% & 1.04\% \\
MCTS-AHD
& 1.92\% & 0.58\% & 1.08\% & 0.38\% & 0.88\% & 0.37\% & 0.87\% \\
PRS (ours)
& \textbf{1.82\%} & \textbf{0.18\%} & \textbf{0.86\%} & 0.35\%
& \textbf{0.63\%} & 0.45\% & \cellcolor{gray!20}\textbf{0.72\%} \\
\noalign{\hrule height 1.2pt}
\end{tabular}
\vspace{-2mm}
\end{table}
\begin{wraptable}{r}{0.55\textwidth}
\vspace{-0.25em}
\footnotesize
\centering
\caption{
Performance of step-by-step constructive heuristics on KP. Near-optimal objectives are obtained by OR-Tools. 
The best-performing method is
shaded, and each test set’s overall best result is in bold.
}
\label{tab:kp_single}

\setlength{\tabcolsep}{1.4pt}
\renewcommand{\arraystretch}{1.08}

\begin{tabular}{l|cc|cc|cc}
\noalign{\hrule height 1.2pt}
\multicolumn{7}{c}{\textbf{Knapsack Problem (KP)}} \\
\hline
\multicolumn{1}{c|}{Test Sets}
& \multicolumn{2}{c|}{N=50}
& \multicolumn{2}{c|}{\underline{N=100}}
& \multicolumn{2}{c}{N=200} \\
\hline
\multicolumn{1}{c|}{Method}
& Obj.$\uparrow$ & Gap$\downarrow$
& Obj.$\uparrow$ & Gap$\downarrow$
& Obj.$\uparrow$ & Gap$\downarrow$ \\
\hline
Optimal
& 20.037 & --
& 40.271 & --
& 57.448 & -- \\
GC
& 19.830 & 1.04\%
& 40.030 & 0.60\%
& 57.115 & 0.58\% \\
POMO
& 19.612 & 2.12\%
& 39.676 & 1.48\%
& 57.271 & 0.31\% \\
\hline
\multicolumn{7}{c}{LLM-based AHD (GPT-5-nano)} \\
\hline
FunSearch
& 19.988 & 0.24\%
& 40.227 & 0.11\%
& 57.398 & 0.09\% \\
EoH
& 19.994 & 0.22\%
& 40.233 & 0.10\%
& 57.398 & 0.09\% \\
MCTS-AHD
& 20.010 & 0.13\%
& 40.250 & 0.05\%
& 57.422 & 0.04\% \\
Ours
& \cellcolor{gray!20}\textbf{20.023} & \cellcolor{gray!20}\textbf{0.07\%}
& \cellcolor{gray!20}\textbf{40.262} & \cellcolor{gray!20}\textbf{0.02\%}
& \cellcolor{gray!20}\textbf{57.436} & \cellcolor{gray!20}\textbf{0.02\%} \\
\noalign{\hrule height 1.2pt}
\end{tabular}
\vspace{-1.2em}
\end{wraptable}
  For OBPP, items arrive sequentially and must be assigned immediately; the heuristic maps the incoming item size together with current bin states (e.g., remaining capacities) to priority scores over bins, placing each item into the best feasible bin. Specifically, Online-BPP instances are generated following EoH using the standard Weibull distribution \citep{castineiras2012weibull}. Regarding KP, Table \ref{tab:kp_single} shows that PRS consistently achieves the lowest or near-lowest optimality gaps across all problem sizes and clearly outperforms existing LLM-based AHD baselines under GPT-5-nano. Table \ref{tab:obpp} further shows that online BPP heuristics designed by PRS have a consistently superior average performance across six test settings. 

\subsection{Comparison on FSSP} 

GLS acts as a perturbation-and-improvement framework, which iteratively perturbs a current schedule using neighborhood operators (e.g., swap and relocate) and employs a heuristic matrix to guide which parts to penalize or prioritize for better exploration \citep{arnold2019knowledge}. Table \ref{tab:full_taillard_tab} reports relative improvements in the makespan in Taillard instances. PRS exhibits a consistent advantage over other methods in nearly all test sets and on average.



\section{Discussions}
\label{sec:discussion}

\begin{wrapfigure}{r}{0.6\columnwidth}
\vspace{-2mm}
\centering

\begin{minipage}{\linewidth}
  \captionsetup{type=table}
  \centering
  \footnotesize
  \setlength{\tabcolsep}{2pt}
  \renewcommand{\arraystretch}{1}
  \caption{Ablation study of PRS on key mechanisms and operators. Results are reported as mean $\pm$ standard deviation over five independent runs.}
  \label{tab:ablation_PRS}

  \begin{tabular}{lcc}
  \specialrule{1.2pt}{0pt}{0pt}
  Methods & CVRP-50$\downarrow$ & KP-100$\uparrow$ \\
  \midrule

    PRS (default setting) 
   & \mstd{\textbf{8.9583}}{0.0726}
   & \mstd{\textbf{40.2603}}{0.0165}\\
     \hline
  w/o Expansion
  & \mstd{9.7492}{0.2840}
  & \mstd{40.2307}{0.0001} \\
  w/o Compression
  & \mstd{9.3737}{0.0707}
  & \mstd{40.2321}{0.0015} \\
  w/o Steering
  & \mstd{9.0728}{0.1207}
  & \mstd{40.2581}{0.0754} \\
  w/o UCB Selection
  & \mstd{9.3781}{0.1142}
  & \mstd{40.2483}{0.0336} \\
    $C=10^5$
  & \mstd{9.1399}{0.1106}
  & \mstd{40.2432}{0.0137} \\
  \hline
  EC Operators (M1\&M2)
  & \mstd{9.2277}{0.0570}
  & \mstd{40.2516}{0.0071} \\

  \specialrule{1.2pt}{0pt}{0pt}
  \end{tabular}
\end{minipage}

\vspace{2mm}

\begin{minipage}{\linewidth}
  \centering
  \begin{subfigure}[t]{0.49\linewidth}
    \centering
    \includegraphics[width=\linewidth]{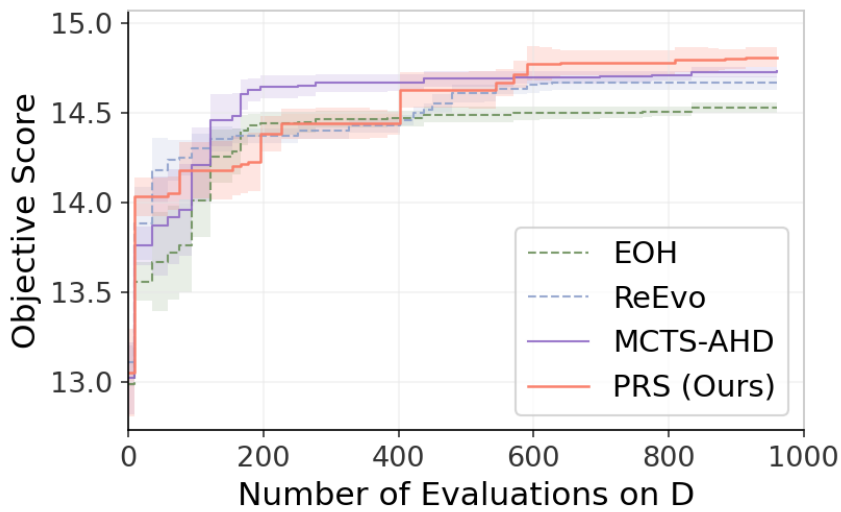}
    \caption{OP.}
    \label{fig:trend-op}
  \end{subfigure}\hfill
  \begin{subfigure}[t]{0.49\linewidth}
    \centering
    \includegraphics[width=\linewidth]{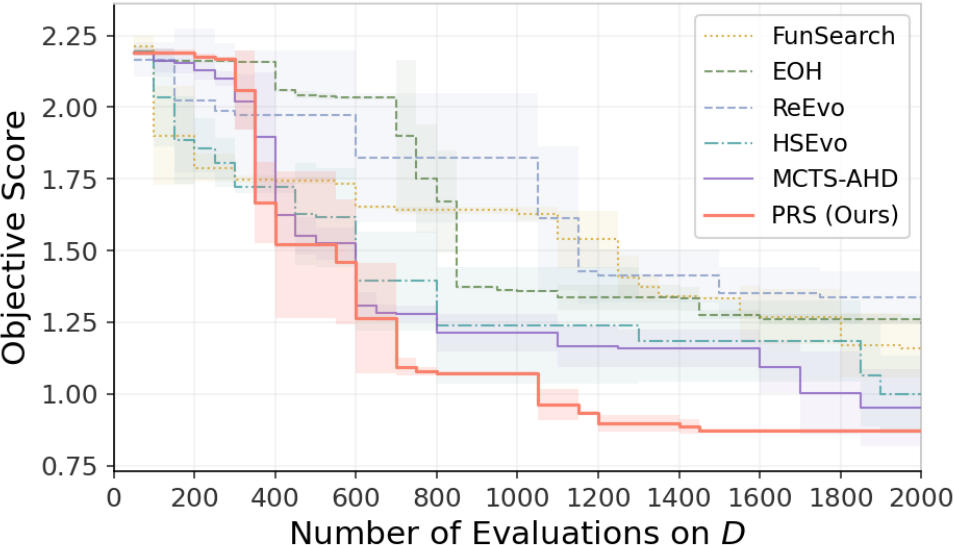}
    \caption{OBPP.}
    \label{fig:trend-fssp}
  \end{subfigure}

  \caption{Evolution curves of the best heuristic in training on two scenarios. Each curve is averaged over five independent runs and the standard deviation is shaded.}
  \label{fig:trend-two}
\end{minipage}

\vspace{-4mm}
\end{wrapfigure}

\textbf{Ablations on Mechanisms.}  We analyze the necessity of each component in PRS. As shown in Table \ref{tab:ablation_PRS}, all three perturbation operators contribute to search effectiveness: expansion and compression support complementary transformation directions by encouraging richer and simpler heuristic designs, respectively, while steering provides fine-grained exploitation around promising heuristic structures. Disabling UCB-based operator selection and replacing it with random selection consistently degrades performance, indicating that adaptively allocating search effort among the three operators is beneficial.  
Disabling the stagnation-aware restart mechanism ($C=10^5$) also weakens performance, further showing the importance of adaptive search control for sustained exploration. Beyond component removal, replacing expansion and compression with EC-style operators (M1\&M2) degrades performance within PRS, suggesting that explicitly targeting complementary structural regimes is more effective than the default EC-style transformations in this setting. 

\textbf{Evolution Curves of Different AHD methods.}
In Fig. \ref{fig:trend-two}, we compare the search dynamics of PRS with representative AHD methods on two representative tasks: OP with ACO and OBPP with the constructive framework. Most baselines exhibit rapid early improvements followed by early convergence, indicating a limited ability to sustain effective exploration as the search progresses. In contrast, PRS shows smoother and more continuous performance gains throughout the search.
This behavior can be attributed to its multi-directional perturbation mechanism, which actively diversifies heuristic generation and mitigates early homogenization, together with the stagnation-aware controller that restores exploration when progress slows. Empirically, the evolution curves suggest that sustained search effectiveness can be achieved through explicitly controlling diversification through directional search primitives without relying on elaborate EC-based operator orchestration.

\textbf{Additional Experimental Results.}
We provide additional experiments in Appendix~\ref{sec:extra_experiments}--\ref{sec:sensitivity_analysis}. In particular, we first provide diversity analyses at both the idea and implementation levels to examine whether the proposed directional search primitives indeed broaden heuristic-space exploration. We then include statistical significance tests (p-values), LLM cost analysis, and sensitivity analysis on the key hyperparameters. Importantly, we further examine how different perturbation directions relate to ID and OOD performance,
and an analysis on the contributions of each directional search operator on different problem settings. Moreover, we provide further experimental results on the TSP and TSPLib~\citep{reinelt1991tsplib} setting, black-box settings, and benchmarking across different LLMs.

\section{Conclusion}
\label{sec:conclusion}
In this paper, we propose PRS, a multi-directional search framework for automatic heuristic design that makes directional heuristic transformation an explicit generation-level search primitive. By organizing heuristic generation through complementary transformation directions and combining them with breadth-first search and adaptive operator control, PRS actively directs diversification, sustains exploration, and mitigates premature convergence. Extensive experiments across diverse combinatorial optimization tasks demonstrate that PRS consistently discovers high-quality heuristics and outperforms existing LLM-based AHD methods in various settings, emphasizing the importance of exploring directionally diverse regions of the heuristic search space.

\bibliography{iclr2027_conference}
\bibliographystyle{iclr2027_conference}

\appendix
\clearpage
\startcontents[appendix]

\section*{Appendix Contents}
\setcounter{tocdepth}{2}
\printcontents[appendix]{}{1}{}

\clearpage

\section{Related Work}
\label{sec:related work}
\subsection{Automated Heuristic Design}

Automated Heuristic Design (AHD), often referred to as hyper-heuristics \citep{burke2018classification}, aims to automatically discover high-performing heuristics from a large space of valid heuristic candidates rather than directly solving problem instances. Most AHD methods adopt evolutionary algorithms as the underlying search strategy to iteratively refine heuristics automatically through selection and generation, as well as a series of defined evolutionary operators for combination and adaptation \citep{pillay2018hyper}. Among them, genetic programming (GP) \citep{langdon2013foundations} has emerged as a prevailing approach for heuristic generation due to its flexibility and interpretability, enabling the synthesis of reusable or instance-specific heuristics via tree-based representations. However, GP-based AHD typically relies on carefully hand-crafted mutation and crossover operators, whose effectiveness is highly sensitive to representation choices and domain knowledge \citep{duflo2019gp}. Beyond GP, recent AHD frameworks further categorize heuristic generation into constructive and perturbative paradigms, or integrate multiple operators and stages through component-based unified solvers, with the goal of enhancing generality across diverse optimization problems \citep{bezerra2015automatic,qu2020general}. 
Despite these advances, existing AHD approaches largely depend on manually designed components and expert knowledge \citep{liu2025eoh}, which remain a major bottleneck for scalability and portability.

\subsection{LLM-based AHD}

Recent work has introduced large language models into automated heuristic design, leading to a class of population-based LLM-driven AHD methods that largely follow an evolutionary paradigm \citep{liu2023algorithm,romera2024mathematical,liu2024evolution,ye2024reevo,liu2024llm4ad}. In general, these methods iterate over three core steps after initially sampling a set of heuristic functions: 
selecting mutation or crossover operators together with parent heuristics from the population, prompting LLMs to generate new heuristic implementations—typically in executable code form—and updating the population (or the tree) based on empirical performance evaluated on a validation dataset until a fixed maximum evaluation step is reached. Representative frameworks within this paradigm include FunSearch \citep{romera2024mathematical}, which adopts a multi-island evolutionary structure with a fixed prompt strategy for algorithm refinement, EoH \citep{liu2024evolution}, which expands the search space through multiple prompt strategies that evolve both intermediate reasoning and code, and ReEvo \citep{ye2024reevo}, which incorporates short- and long-term reflection mechanisms to guide heuristic evolution. More recently, diversity has received increasing attention as an important factor for sustaining exploration in LLM-based AHD. For example, HSEvo~\citep{dat2025hsevo} explicitly promotes population diversity through diversity-aware search and population management, showing the value of preserving heterogeneous heuristic candidates during evolution.

Building on this direction, recent studies have introduced more explicit mechanisms for measuring diversity and structuring the search space. BehaveSim~\citep{zhang2026rethinking} measures algorithmic similarity from execution trajectories and incorporates behavioral diversity into candidate selection. PartEvo~\citep{hu2025partition} projects candidate algorithms into feature spaces and constructs semantic niches to allocate search effort across different regions of the abstract language space. ShinkaEvolve~\citep{lange2025shinka} further promotes exploration through archive-based parent sampling and code-novelty rejection based on embedding similarity. These approaches demonstrate the importance of maintaining diverse candidates during LLM-based algorithm search, but primarily regulate diversity through similarity assessment, population organization, novelty filtering, or search-space partitioning. In contrast, PRS controls diversification directly at the generation level by explicitly selecting how a parent heuristic is transformed before realization, using expansion, compression, and steering as complementary directional perturbation operators.



\subsection{LLM for CO}
Beyond LLM-based AHD, the use of LLMs for tackling combinatorial optimization (CO) problems has also emerged as a promising direction, and one line of research treats LLMs as optimizers. Starting with OPRO \citep{yang2023large} that iteratively refines candidate solutions for TSP by conditioning on previously generated solutions and their objective values, the existing LLM-as-optimizer methods \citep{guo2023towards,liu2024large,elhenawy2024visual} improve solution quality through repeated in-context prompting, evaluation, and reinjection, forming a closed-loop optimization process. Another closely related direction employs LLMs as end-to-end solvers (LLM-as-E2E-solver) that directly map problem descriptions to complete solutions, either via prompting or with additional fine-tuning within task-specific environments \citep{abgaryan2024llms,jiang2025large}. These approaches demonstrate the potential of LLMs to reason over complicated CO structures and to generate feasible solutions without explicit algorithmic decomposition and solver utilization.

\subsection{Adaptive Control in EC and LLM-based AHD.}
Adapting search strategies based on the evolving state has long been a crucial objective in evolutionary computation. Traditional approaches mainly rely on low-level mechanisms such as Adaptive Operator Selection (AOS) \citep{fialho2010adaptive} and parameter control \citep{eiben2015evolutionary}, which adjust mutation rates or operator probabilities in a reactive manner. Such approaches operate through numerical credit assignment mechanisms that bias the search toward recently successful strategies, but they lack the capacity to reason about search dynamics, for example, detecting stagnation or loss of diversity. More recent LLM-based AHD methods have also introduced higher-level mechanisms for adapting search behavior, such as diversity-aware population management and stagnation handling \citep{dat2025hsevo}. These developments motivate adaptive search control beyond fixed operator schedules. Driven by these motivations, PRS combines adaptive operator control with stagnation-aware restart, where operator selection is continuously adjusted according to recent search utility, while prolonged stagnation triggers an elite restart and temporary reheating of exploration.

\section{Evaluated Problem Settings}
\label{sec:problem_settings}
\subsection{Definition of Tasks}
\label{sec:definition of tasks}

We evaluate the proposed PRS framework on a diverse set of combinatorial
optimization problems, covering routing, packing, scheduling, and combinatorial set construction tasks. Specifically, they are TSP, CVRP, OP, MKP, BPP, KP, ASP and FSSP. For BPP, we only test its online setting (online BPP). In this subsection, we will introduce the mathematical formulation and evaluation protocols of these problems in detail.

\paragraph{Traveling Salesman Problem (TSP).}
Given a set of $n$ cities with pairwise distances, the Traveling Salesman Problem
aims to find the minimum-length tour that visits each city exactly once and returns
to the starting city.

\begin{equation}
\min_{\pi\in S_n}\;\sum_{t=1}^{n} d_{\pi_t,\pi_{t+1}},\quad \pi_{n+1}=\pi_1,
\label{eq:tsp}
\end{equation}
where $d_{ij}$ is the Euclidean distance between city $i$ and city $j$, $\pi_t$ is the selected node at step $t$ of a certain valid tour (i.e., a Hamiltonian tour over $n$ cities) and $S_n$ is the solution space of the TSP with $n$ cities. We consider the symmetric TSP formulation and evaluate heuristics on both synthetic
instances and real-world benchmarks from TSPLib.
Following ReEvo, synthetic TSP instances sample nodes uniformly from the unit square
$[0,1]^2$ \citep{ye2024reevo}.

\paragraph{Capacitated Vehicle Routing Problem (CVRP).}
CVRP consists of a depot indexed 0 and a set of customers $\{1,\dots,n\}$, each customer $i$ has a demand $q_i$ (with $q_0 = 0$), and all vehicles have an identical capacity $Q$.
The objective is to design a set of closed routes that serve all customers without
exceeding vehicle capacities, while minimizing the total travel cost:
\begin{equation}
\min \sum_{k} \sum_{(i,j)\in \pi_k} d_{ij}
\end{equation}
subject to
\begin{equation}
\sum_{i \in \pi_k} q_i \le Q, \quad \forall k.
\end{equation}
where each sub-route $\pi_k$ starts and ends at the depot.
\paragraph{Orienteering Problem (OP).}
Given a graph with nodes $\{0,1,\dots,n\}$, where node $0$ denotes the depot and is associated with zero reward ($r_0=0$), each remaining node $i$ has a reward $r_i$, and $d_{ij}$ denotes the travel distance.
The goal of OP is to find a route $\pi = (\pi_1,\dots,\pi_k)$ starting and ending at the depot
that maximizes the total reward collected under a travel-length budget $B$:
\begin{equation}
\max_{\pi} \sum_{i \in \pi} r_i
\quad \text{s.t.} \quad
\sum_{t=0}^{k} d_{\pi_t,\pi_{t+1}} \le B
\end{equation}
where $\pi_0 = \pi_{k+1} = 0$ and each node is visited at most once.
\paragraph{Knapsack Problems (KP and MKP).}
The Knapsack Problem (KP) aims to select a subset of items to maximize total profit
subject to a capacity constraint.
Given item values $\{v_i\}$, item weights $\{w_i\}$, and knapsack capacity $C$, KP aims to
\begin{equation}
\max_{x\in\{0,1\}^n}\;\sum_{i=1}^n v_i x_i \quad \text{s.t.}\quad \sum_{i=1}^n w_i x_i \le C.
\label{eq:kp}
\end{equation}
where $x_i$ is a binary indicator of whether the item $i$ is packed into the knapsack or not. We follow AM \citep{kool2018attention} in generating instances and take $W = 25$ for 100-item and 200-item KP instances and
$W = 12.5$ for 50-item ones.

The Multiple Knapsack Problem (MKP) extends this formulation by introducing
multiple resource constraints.
We evaluate heuristics under a constructive selection setting, where items are
considered sequentially. For the Multiple knapsack problem (MKP), we follow the instance generation
procedure in ReEvo \citep{ye2024reevo}. Specifically, item values and weights are
independently sampled from a uniform distribution on $[0,1]$, and the capacity of
each knapsack is set proportionally to the total item weights to ensure non-trivial
feasible solutions.
 Specifically for the knapsack problem, \texttt{select\_next\_item} function takes as input the remaining capacity, a numpy array of item values, and a numpy array of item weights and returns the next item to pick.
For both KP and MKP, feasibility is enforced at each step, and the total profit of
the selected items serves as the objective value.

\paragraph{Online Bin Packing Problem (Online BPP).}
In the Online Bin Packing Problem, items arrive sequentially and must be assigned
to bins immediately upon arrival without knowledge of future items. Given the bin capacity $W$ and an item stream $\{s_t\}$, the objective is to minimize the number of bins used.
Formally, upon receiving $s_t$, the policy must choose a bin index $b_t$ with the highest priority score. We follow FunSearch \citep{romera2024mathematical} and EoH \citep{liu2024evolution} to
generate Weibull instances for online BPP.

\paragraph{Flow Shop Scheduling Problems (FSSP).}
Scheduling problems aim to determine the execution order of operations to optimize
the overall schedule length.
In the Flow Shop Scheduling Problem (FSSP), $n$ jobs must be processed on $m$ machines,
where each job consists of $m$ operations executed in the same predetermined order across all machines.
Each machine can process at most one operation at a time, and the objective is to minimize the makespan, i.e., the total schedule length.

Following prior work \citep{emmons2012flow}, we consider a perturbation-based
scheduling setting, where the processing order remains consistent during each step, and the machines cannot execute multiple operations simultaneously.
During heuristic search, we perform optimization on a set of randomly generated
instances $D$.
Each instance in $D$ consists of 50 jobs and between 2 and 20 machines, with processing
times sampled independently from a uniform distribution over $[0,1]$ \citep{pan2021deep}.
The average makespan over these instances is used as the fitness value for heuristic
evaluation. 

\paragraph{Admissible Set Problem (ASP).}
The admissible set problem (ASP) aims to construct the largest possible set of vectors
$A \subset \{0,1,2\}^n$,
where each vector has exactly $w$ non-zero elements.
Formally, ASP can be written as
\begin{equation}
\max |A|
\end{equation}
subject to
\begin{equation}
\sum_{i=1}^{n}\mathbb{I}[a_i\neq 0]=w,\quad \forall a\in A,
\end{equation}
and
\begin{equation}
\exists i\in\{1,\dots,n\}\ \text{s.t.}\ \{a_i,b_i,c_i\}\in
\bigl\{\{0,0,1\},\{0,0,2\},\{0,1,2\}\bigr\},
\quad \forall a,b,c\in A,\ a\neq b\neq c.
\end{equation}

Under the step-by-step constructive framework, the heuristic to be designed is a priority function that assigns a score to each candidate vector according to its suitability for being added to the current admissible set. Higher scores indicate higher priority.

\subsection{Evaluated Framework for Tasks}
\label{sec: evaluated framework}
We evaluate PRS on eight NP-hard combinatorial optimization problems (COPs) under three widely used algorithmic frameworks for automatic heuristic design (AHD).
Specifically, we consider (i) \textbf{step-by-step construction} for TSP, KP, ASP and Online-BPP;
(ii) \textbf{ACO} for MKP, CVRP and OP; and (iii) \textbf{GLS} for TSP and FSSP.

\paragraph{Evaluation metric.}
For minimization problems (TSP, CVRP, Online-BPP, FSSP), we report the relative optimality gap:
\begin{equation}
\mathrm{Gap}(\%) \;=\; \frac{f(h)-f^\star}{f^\star}\times 100\%,
\label{eq:gap_min}
\end{equation}
where $f(h)$ is the objective value returned by a heuristic function $h$ and $f^\star$ is the optimal reference value
(e.g., best-known or a lower bound). For maximization problems (OP, KP, MKP, ASP), we use
\begin{equation}
\mathrm{Gap}(\%) \;=\; \frac{f^\star-f(h)}{f^\star}\times 100\%.
\label{eq:gap_max}
\end{equation}
where $f^\star$ is the best-known value or an upper bound.

\subsubsection{Step-by-step Constructive Framework}
Step-by-step construction is a generic framework for combinatorial optimization that incrementally builds a solution by repeatedly selecting and adding the highest-priority candidate until a complete feasible solution is obtained. Nearest-neighbor is a widely used handcrafted step-by-step constructive heuristic for TSP, which selects the next node solely based on its distance to the current node. 
Analogously, for the knapsack problem, a greedy heuristic is adopted that prioritizes items according to their value-to-weight ratio. 
We refer to these two manually designed heuristics as the Greedy Construct (GC) baseline in Table \ref{tab:kp_single} and Table \ref{tab:tsp_step_by_step}.

\begin{itemize}
    \item \textbf{TSP.} Under the step-by-step construction framework, the designed heuristic is a
\texttt{select\_next\_node} function that scores all unvisited candidates and greedily appends the best one. For each instance, the heuristic outputs the best next city to select based on the current node, the destination node (starting node), the indices of all unvisited nodes and the symmetric distance matrix. 
    \medskip
    \item \textbf{KP.} In step-by-step construction for KP, the heuristic scores the remaining items given $(v_i,w_i)$ and the remaining capacity,
then selects the next item to pack.
    \medskip
    \item \textbf{Online BPP. }We evaluate heuristics under this strict online setting, where the heuristic
determines the bin assignment for each incoming item based solely on current bin
states and the current item's size.
\medskip
\item \textbf{ASP.} In step-by-step construction for ASP, the heuristic scores each candidate vector given the vector itself, the problem dimension $n$, and the weight parameter $w$, and then selects high-priority vectors to incrementally build the admissible set while preserving feasibility.
\end{itemize}

\subsubsection{GLS-based Problems}
\paragraph{Guided Local Search (GLS).}
For GLS-based evaluation, we embed the LLM-designed heuristic into a standard GLS loop (e.g., maintaining a penalty-based edge weights matrix). We use two local operators (i.e., Swap and
Relocate) for perturbation and local improvements, and evaluate after each perturbation-local search cycle.

\paragraph{TSP under GLS.}
We apply GLS to TSP by optimizing an edge-weight matrix that penalizes frequently used edges. Let $c_{ij}$ be
the edge cost between node $i$ and node $j$ and let $p_{ij}\in\mathbb{Z}_{\ge 0}$ be a penalty. The augmented edge cost takes the typical form:
\begin{equation}
c'_{ij} \;=\; c_{ij} + \lambda\, p_{ij},
\label{eq:gls_aug}
\end{equation}
where $\lambda$ controls penalty strength. The AHD target is improving the penalty (utility) heuristic (e.g., how to update
$p_{ij}$). We apply LLM-based AHD methods to design the core heuristic for generating
knowledge matrices within the knowledge-guided local search (KGLS) framework
\citep{arnold2019knowledge}. For TSP, KGLS maintains a current solution together with the
best solution encountered so far. At each iteration, the current solution is
first perturbed according to the learned knowledge matrix, followed by the local
search using 2-opt and relocate operators \citep{sengupta2019local,tuononen2022analysis}.
We use scale-wise GLS budgets summarized in Table~\ref{tab:gls_params}.

\paragraph{Flow Shop Scheduling Problem (FSSP) under GLS.} We evaluate FSSP within a GLS-style perturbation and local improvement procedure, where the evolved heuristic
controls (i) which operations to penalize (or prioritize) and (ii) which parts of the schedule to perturb.

\subsubsection{ACO-based Problems}
\label{subsec:aco}

For ACO-based problems, including CVRP and OP, we adopt a standard ant colony optimization (ACO)
framework that mimics the nature of ants to find routes between food sources and the colony.
In this setting, the LLM-designed component is the heuristic information function, which defines the edge desirability matrix $\eta$ used by ants during solution construction.

At each construction step, an ant located at node $i$ selects the next node $j$
according to the transition probability
\begin{equation}
P_{ij} \;=\;
\frac{\tau_{ij}^{\alpha}\,\eta_{ij}^{\beta}}
{\sum_{k \in \mathcal{A}(i)} \tau_{ik}^{\alpha}\,\eta_{ik}^{\beta}},
\label{eq:aco_transition}
\end{equation}
where $\tau_{ij}$ denotes the pheromone value, $\eta_{ij}$ is the heuristic information
produced by the evolved heuristics, $\alpha$ and $\beta$ are fixed hyperparameters,
and $\mathcal{A}(i)$ denotes the set of feasible next nodes under the problem-specific constraints \citep{dorigo1992optimization,dorigo2006ant}.

The heuristic matrix functions under ACO framework have different semantics for CVRP, OP, and MKP:
\begin{itemize}
    \item \textbf{CVRP.} Under the ACO framework, ants sequentially construct routes by selecting the next customer based on pheromone trails and LLM-generated heuristic scores.
    \item \textbf{OP.} Within the ACO framework, ants incrementally build candidate paths by choosing the next site according to the pheromone information and LLM-based heuristics that balance reward and travel budget.
     \item \textbf{MKP.} For MKP, the ACO framework is applied in a black-box manner, which ants stochastically sample item selections guided by pheromone information and LLM-generated heuristic scores that indicate the desirability of including each item in the solution.
\end{itemize}
The pheromone update rule and other ACO hyperparameters are kept identical across all methods to ensure a fair comparison.
Only the heuristic information function $\eta$ is replaced by the LLM-designed heuristics generated by PRS or the baseline AHD.

\clearpage
\section{Complete Algorithm}
\label{sec:full algo}
Algorithm 1 provides the complete pseudo-code for the PRS framework.
The full framework includes an initialization stage (up to line 6) and an iterative stage consisting of a BFS-based search process with multi-directional perturbation operators and stagnation-aware mechanisms.

\begin{algorithm}[htbp]
\caption{\textbf{PRS}: Perturbation--Rewrite Search via BFS and Adaptive Control}
\label{alg:PRS}
\begin{algorithmic}[1]
\STATE \textbf{Input:}
Validation dataset $D$; evaluation environment $g$;
heuristic pool size $N$; initial number of sampled heuristics $N_{\mathrm{init}}$; maximum
evaluation steps $T$;
root selection via $\epsilon$-greedy;
BFS parameters: depth $d$, beam size $\kappa$, frontier size $fs$, softmax temperature $T_{\mathrm{sel}}$;
UCB operator selection with exploration coefficients $c_0$ and $c_T$;
stagnation trigger threshold $\mathcal{C}$ and reheating horizon $\Delta=0.1T$;
perturbation operators $\Phi_\theta$ and rewrite operator $\Psi_\theta$.

\STATE Get $g_0=g(h_{\mathrm{seed}})$ where $h_{\mathrm{seed}}$ is the seed heuristic function and set $t\leftarrow1$
\STATE Initialize candidate set $\mathcal{P} \gets \{h_{\mathrm{seed}}\} \cup\{h_1,\ldots,h_{N_{\mathrm{init}}-1}\}$ using $h_{\mathrm{seed}}$ and an initialization prompt
\STATE Evaluate all newly generated $h\in\mathcal{P}$ using $g$
\STATE $h^\star \leftarrow \arg\max_{h\in\mathcal P} g(h)$; \ $t \gets |\mathcal{P}|$
\STATE $\mathcal{P}\leftarrow\pi_{\mathrm{greedy}}(\mathcal{P},N)$; initialize operator reward window $\mathcal{H}$; $t'\leftarrow-\infty$

\WHILE{$t < T$}

    \STATE $c_t^{\mathrm{base}} \leftarrow c_0+(c_T-c_0)\frac{t}{T}$

    \IF{$t-t' < \Delta$}
        \STATE $\hat{t}\leftarrow\mathrm{clip}\!\left(\frac{t-t'}{\Delta},0,1\right)$
        \STATE $c_t \leftarrow c_t^{\mathrm{base}}+
        (c_0-c_t^{\mathrm{base}})(1-\hat{t})$
        \hfill // operator reheating
    \ELSE
        \STATE $c_t\leftarrow c_t^{\mathrm{base}}$
    \ENDIF

    \STATE $h_{\mathrm{root}}\leftarrow\mathrm{EpsilonGreedy}(\mathcal{P},\epsilon)$

    \STATE $\mathcal{F}\leftarrow\{h_{\mathrm{root}}\}$ \hfill // BFS frontier (layer 0)
    \STATE $\mathcal{S}\leftarrow\emptyset$

    \FOR{$\ell=1$ \textbf{to} $d$}
        \STATE $\mathcal{C}_{\ell}\leftarrow\emptyset$

        \FOR{each parent $h\in\mathcal{F}$}
            \STATE Select $\Phi_\theta\in
            \{\Phi^{\mathrm{expn}}_\theta,\Phi^{\mathrm{comp}}_\theta,\Phi^{\mathrm{steer}}_\theta\}$
            using UCB with coefficient $c_t$
            \STATE $\tilde{z}\leftarrow\Phi_\theta(z_h\mid\mathcal{C})$
            \hfill // perturbation
            \STATE $\{h^{(1)},\ldots,h^{(\kappa)}\}
            \sim\Psi_\theta(h,\tilde{z},\Phi_\theta\mid\mathcal{C})$
            \hfill // rewrite
            \STATE Evaluate all $h^{(j)}$ using $g$; $t\leftarrow t+\kappa$
            \STATE Update $\mathcal{H}$ using the best offspring improvement;
            $\{\mathcal{C}_\ell,\mathcal{S}\}\leftarrow
            \{\mathcal{C}_\ell,\mathcal{S}\}\cup\{h^{(j)}\}_{j:\mathrm{valid}}$

            \IF{$t\ge T$}
                \STATE \textbf{break}
            \ENDIF
        \ENDFOR

        \IF{$t\ge T$ or $\mathcal{C}_\ell=\emptyset$}
            \STATE \textbf{break}
        \ENDIF

        \STATE $\mathcal{F}\sim\mathrm{Sample}(\mathcal{C}_\ell,p(h_i))$
        \hfill // According to Eq.~(\ref{eq:softmax_frontier})
    \ENDFOR

    \STATE $\mathcal{P}\leftarrow\pi_{\mathrm{sel}}(\mathcal{P}\cup\mathcal{F})$
    \hfill // preserve best and randomly retain remaining candidates
    \STATE Update $h^\star$ using all evaluated offspring in $\mathcal{S}$

    \IF{no global-best improvement for $\mathcal{C}$ evaluations}
        \STATE $\mathcal{P}\leftarrow\{h^\star\}$; $t'\leftarrow t$; clear $\mathcal{H}$
        \hfill // Elite Pool Reset + Operator Reheating
    \ENDIF
\ENDWHILE

\STATE \textbf{return} Best valid heuristic $h^\star$ found

\end{algorithmic}
\end{algorithm}

\clearpage

\section{Detailed Methodology}
\label{sec:detailed methodology}
\subsection{PRS Prompt}
\label{sec:app_propmt}
PRS uses a shared prompt library across all problem settings. It contains two system prompts that regulate the behavior of the LLM as an idea generator and as a code generator, respectively, as well as seven workflow prompt templates. These templates are used for: (i) describing the seed heuristic as a natural-language idea, (ii) generating initial heuristic ideas, (iii) applying the Expansion operator, (iv) applying the Compression operator, (v) applying the Steering operator, (vi) realizing an initial idea as executable code, and (vii) rewriting a parent implementation to realize a target idea.

 Besides those common prompts, each problem setting contains two types of problem-specific prompts: function contract which dictates the format (including the names and types of inputs and outputs) and robustness constraints of designed heuristics, and a task description prompt (i.e., problem description). In this section, we describe their meaning and prompt engineering in detail, where we highlight the uniqueness of the prompt design in red colors. For problem-specific prompt components, we take the problem setting Knapsack Problem in step-by-step constructive framework as an example. 
 


\definecolor{PromptGreen}{RGB}{0,130,0}     
\definecolor{PromptYellow}{RGB}{180,120,0}  
\definecolor{PromptPurple}{RGB}{120,60,160} 
\definecolor{PromptRed}{RGB}{160,40,40}     
\definecolor{PromptBlue}{RGB}{20,30,80}
\definecolor{PromptGray}{gray}{0.45}
\definecolor{PromptGreen}{RGB}{0,130,0}
\definecolor{SystemGray}{RGB}{60,60,60}

\definecolor{offspringblue}{RGB}{120,189,152}
\tcbset{
  basepromptbox/.style={
    enhanced,
    width=0.95\linewidth,
    center,
    colframe=PromptBlue,
    colbacktitle=PromptBlue,
    colback=gray!5,
    fonttitle=\sffamily\bfseries\normalsize,
    toptitle=4pt,
    bottomtitle=4pt,
    arc=6pt,
    boxrule=1.2pt,
    left=6pt,
    right=6pt,
    top=6pt,
    bottom=6pt,
  }
}












\begin{itemize}
  \item \textbf{Task Description and Function Contract:} While the seed heuristics are only used once in the heuristic pool initialization prompt, the task description and function contract are employed not only during initialization, but also throughout the perturbation and rewrite stages of the evolutionary process. The robustness constraints in function contract do not add inductive bias that explicitly improves the performance of heuristics. Instead, they ensure the validity and consistency of the designed heuristics (e.g., the return must be of the same shape as the distance matrix). Below, we provide examples of the task description and function contract.
\end{itemize}

\newtcolorbox{taskdescpromptbox}[1][]{%
  basepromptbox,
  title=Task Description Prompt,
  #1
}
\begin{taskdescpromptbox}

Solving 0-1 Knapsack Problem (KP) with constructive heuristics.
The goal is to incrementally build a feasible solution by selecting items one by one,
so as to maximize the total value under a weight (capacity) constraint. Each item can be selected at most once.
At each step, the heuristic observes the current remaining capacity and the full list of item weights and values,
and decides which item to select next. The selection is irreversible and directly affects future decisions.

\end{taskdescpromptbox}
\newtcolorbox{functioncontractpromptbox}[1][]{%
  basepromptbox,
  title=Function Contract Prompt,
  #1
}
\begin{functioncontractpromptbox}
You must implement the function:

def \textcolor{PromptYellow}{select\_next\_item}(remaining\_capacity, weights, values):

\ \ \ \ return next\_item

\medskip

inputs:

 -\textcolor{PromptGreen}{remaining\_capacity}: int\\ -\textcolor{PromptGreen}{weights}: np.ndarray \\ -\textcolor{PromptGreen}{values}: np.ndarray

outputs: 

-\textcolor{PromptGreen}{next\_item}: int

\medskip

Robustness constraints:

- The \textcolor{PromptYellow}{select\_next\_item} function takes as input the remaining capacity, a numpy array of item values, and a numpy array of item weights, 
and returns the next item to pick.

\end{functioncontractpromptbox}

\newpage
\begin{itemize}
\item \textbf{System Prompts:} PRS uses two system prompts to separate semantic heuristic reasoning from executable code generation. The \textit{system idea prompt} instructs the LLM to reason at the level of decision principles, interactions, and hyperparameters that affect behaviors, and to produce a concrete natural language heuristic idea. The \textit{system code prompt} instructs the LLM to translate a supplied heuristic idea into a complete executable implementation. This separation allows the three directional search primitives to operate directly on explicit heuristic ideas, while code realization is guided by the resulting target idea and the required function interface.
\end{itemize}

\newtcolorbox{systemideabox}[1][]{%
basepromptbox,
title=System Idea Prompt,
#1
}

\newtcolorbox{systemcodebox}[1][]{%
basepromptbox,
title=System Code Prompt,
#1
}

\begin{systemideabox}
You are an expert in optimization heuristics and algorithm design.
\textcolor{PromptRed}{Your task is to reason at the level of heuristic ideas, decision principles, interactions, and search logic.} Do NOT output Python code.\\
\# Describe the heuristic idea concretely enough that another expert can implement it without guessing important decision logic.\\
\# Clearly explain what information or signals are used, how they affect decisions, how different mechanisms interact, and when the behavior changes.\\
\textcolor{PromptRed}{\# For every behavior-affecting hyperparameter, threshold, schedule, or fallback, state its role and concrete value or rule. Do not leave implementation-critical parameter choices implicit.}\\
\# Focus on algorithmic and heuristic logic rather than Python syntax, variable names, code length, or implementation layout.\\
\# Avoid cosmetic wording changes and parameter-only tweaks unless they meaningfully change the decision mechanism.\\
Return only the heuristic idea description.
\end{systemideabox}

\begin{systemcodebox}
You are an expert in optimization heuristics and Python implementation.
\textcolor{PromptRed}{Your task is to translate the supplied heuristic idea into a complete executable Python heuristic.}\\
\# Return only Python code wrapped in `python ... `.\\
\# Write all necessary imports inside the function when needed.\\
\# Do not output explanations, comments, or docstrings.\\
\# Do NOT use type annotations in the function signature.
\end{systemcodebox}

\newpage
\begin{itemize}
\item \textbf{Initialization Prompts:} PRS initializes the heuristic pool in two stages. The initial idea prompt generates a distinct heuristic idea from the seed idea, and the initial realization prompt translates it into executable code using the seed implementation as context.
\end{itemize}

\newtcolorbox{initialideabox}[1][]{%
basepromptbox,
title=Initial Idea Prompt,
#1
}

\newtcolorbox{initialrealizationbox}[1][]{%
basepromptbox,
title=Initial Realization Prompt,
#1
}

\begin{initialideabox}

\textbf{Task:}\\
\textcolor{PromptGreen}{\texttt{\{task\_desc\}}}

\medskip
\textbf{Function Format:}\\
\textcolor{PromptGreen}{\texttt{\{contract\}}}

\medskip
\textbf{Seed Heuristic Idea:}\\
\textcolor{PromptGreen}{\texttt{\{seed\_idea\}}}

\medskip
\textcolor{red}{\textbf{Propose one distinct heuristic idea for the same task.}}
The proposal must remain compatible with the required function format while introducing a \textcolor{red}{\textbf{meaningful algorithmic change}} rather than cosmetic rewriting.

\medskip
\textbf{Description Requirement:}\\
Describe the proposed heuristic concretely. Explain its main decision principle, what information it uses, how that information influences decisions, how different mechanisms interact, and any conditions under which its behavior changes.

\medskip
\textcolor{red}{The description should be sufficiently specific to guide a separate implementation step without requiring the implementer to invent missing heuristic logic.}

\medskip
\textbf{Output Requirement:}\\
Do not discuss Python syntax or implementation layout. Do not output code.\\
Return only the proposed heuristic idea description.

\end{initialideabox}

\begin{initialrealizationbox}

\textbf{Function Name:}\\
\textcolor{PromptGreen}{\texttt{\{func\_name\}}}

\medskip
\textbf{Task:}\\
\textcolor{PromptGreen}{\texttt{\{task\_desc\}}}

\medskip
\textbf{Function Format:}\\
\textcolor{PromptGreen}{\texttt{\{contract\}}}

\medskip
\textbf{Target Heuristic Idea:}\\
\textcolor{PromptGreen}{\texttt{\{target\_idea\}}}

\medskip
\textbf{Seed Implementation for Context:}\\
\textcolor{PromptGreen}{\texttt{\{seed\_code\}}}

\medskip
\textcolor{red}{\textbf{Implement the TARGET HEURISTIC IDEA as a complete executable heuristic.}}
The target idea is the semantic specification. Use the seed implementation only as low-level implementation context when useful.

\medskip
\textbf{Implementation Requirement:}\\
Faithfully realize the stated decision principles, information signals, interactions, conditions, and adaptive behavior. Do not invent additional heuristic mechanisms that are not implied by the target idea.\\
When low-level implementation details are unspecified, choose simple implementations that preserve the intended decision logic.

\medskip
\textbf{Output Requirement:}\\
Return only valid Python code wrapped in \texttt{```python ... ```}.

\end{initialrealizationbox}
\begin{itemize}
\item \textbf{Multi-directional Idea Operators:} During BFS search, PRS applies one of three idea operators to transform each parent idea into a target idea. \textcolor{red}{Expansion} enriches the heuristic with new decision mechanisms, \textcolor{red}{Compression} removes or consolidates unnecessary mechanisms, and \textcolor{red}{Steering} improves the current strategy through localized modifications. The target idea is subsequently realized by a separate code generation prompt.
\end{itemize}

\newtcolorbox{ideaexpansionbox}[1][]{%
basepromptbox,
title=Expansion Operator Prompt,
#1
}

\newtcolorbox{ideacompressionbox}[1][]{%
basepromptbox,
title=Compression Operator Prompt,
#1
}

\newtcolorbox{ideasteeringbox}[1][]{%
basepromptbox,
title=Steering Operator Prompt,
#1
}

\begin{ideaexpansionbox}

\textbf{Task:}\\
\textcolor{PromptGreen}{\texttt{\{task\_desc\}}}

\medskip
\textbf{Current Heuristic Idea:}\\
\textcolor{PromptGreen}{\texttt{\{parent\_idea\}}}

\medskip
\textcolor{red}{\textbf{Perform IDEA EXPANSION.}}
Develop the current heuristic into a richer algorithmic idea while preserving its useful backbone. Introduce meaningful new decision logic, such as additional information signals, interactions between existing mechanisms, context-dependent rules, or adaptive behavior.

\medskip
\textbf{Expansion Requirement:}\\
Do not simply append independent terms. Explain specifically how every newly introduced mechanism changes the heuristic's decisions and how it interacts with the retained mechanisms.

\medskip
\textcolor{red}{The resulting description must be concrete enough to guide implementation directly: state what information is used, how it influences decisions, how mechanisms are combined, and under what situations the behavior changes.}

\medskip
\textbf{Output Requirement:}\\
Do not increase complexity merely through extra wording, constants, parameters, duplicated terms, or implementation details.\\
Do not output code. Return only the expanded heuristic idea description.

\end{ideaexpansionbox}

\begin{ideacompressionbox}

\textbf{Task:}\\
\textcolor{PromptGreen}{\texttt{\{task\_desc\}}}

\medskip
\textbf{Current Heuristic Idea:}\\
\textcolor{PromptGreen}{\texttt{\{parent\_idea\}}}

\medskip
\textcolor{red}{\textbf{Perform IDEA COMPRESSION.}}
Simplify the current heuristic at the level of algorithmic reasoning while preserving its most useful core principle. Remove, merge, replace, or decouple mechanisms that are redundant, weak, overly specific, or unnecessarily interacting.

\medskip
\textbf{Compression Requirement:}\\
When removing or merging a mechanism, explain how the remaining mechanisms should compensate for its role so that the resulting heuristic remains complete and directly implementable.

\medskip
\textcolor{red}{The resulting description must still specify a complete decision rule: what information is retained, how it influences decisions, how the remaining mechanisms interact, and when behavior changes.}

\medskip
\textbf{Output Requirement:}\\
Compression must reduce conceptual burden rather than merely shorten the description, delete syntax, remove lines, or simplify implementation details.\\
Do not output code. Return only the compressed heuristic idea description.

\end{ideacompressionbox}

\begin{ideasteeringbox}

\textbf{Task:}\\
\textcolor{PromptGreen}{\texttt{\{task\_desc\}}}

\medskip
\textbf{Current Heuristic Idea:}\\
\textcolor{PromptGreen}{\texttt{\{parent\_idea\}}}

\medskip
\textcolor{red}{\textbf{Perform IDEA STEERING.}}
Preserve the core heuristic strategy and its overall conceptual structure, but improve it through localized modifications. Exploit the strengths of the current idea rather than redirecting it toward a fundamentally different heuristic family.

\medskip
\textbf{Steering Requirement:}\\
Focus on fine-grained changes such as adjusting how existing signals are weighted or combined, modifying thresholds, schedules, or parameter settings, refining conditional logic or activation conditions, changing local interactions between existing mechanisms, or strengthening one specific part without altering the overall strategy.

\medskip
\textcolor{red}{Parameter-only changes are allowed only when they induce a meaningful behavioral change. Avoid purely cosmetic rewriting.}

\medskip
\textbf{Output Requirement:}\\
Describe the locally steered heuristic concretely enough for direct implementation: state what local mechanism changes, how it affects decisions, and when it is active.\\
Do not output code. Return only the steered heuristic idea description.

\end{ideasteeringbox}

\begin{itemize}
\item \textbf{Heuristic Rewrite Prompt:} After an idea operator generates a target idea, PRS rewrites the parent implementation to realize that idea as executable code. The prompt receives the target idea, the selected operation, and the parent implementation. The target idea serves as the semantic specification of the offspring, while the operation specifies the intended type of change. The rewrite may modify the parent code as needed, but it must preserve parent mechanisms only when they remain consistent with the target idea.
\end{itemize}

\newtcolorbox{rewritepromptbox}{basepromptbox,title={Heuristic Rewrite Prompt}}

\begin{rewritepromptbox}

\textbf{Function Name:}\\
\textcolor{PromptGreen}{\texttt{\{func\_name\}}}

\medskip
\textbf{Task:}\\
\textcolor{PromptGreen}{\texttt{\{task\_desc\}}}

\medskip
\textbf{Function Format:}\\
\textcolor{PromptGreen}{\texttt{\{contract\}}}

\medskip
\textbf{Target Heuristic Idea:}\\
\textcolor{PromptGreen}{\texttt{\{target\_idea\}}}

\medskip
\textbf{Operation that Produced the Target Idea:}\\
\textcolor{PromptGreen}{\texttt{\{operation\}}}

\medskip
\textbf{Parent Implementation:}\\
\textcolor{PromptGreen}{\texttt{\{parent\_code\}}}

\medskip
\textcolor{red}{\textbf{Rewrite the parent implementation into a complete executable heuristic that faithfully realizes the TARGET HEURISTIC IDEA.}}

\medskip
The target idea is the semantic specification of the new heuristic; the operation states the intended kind of change.

\medskip
\textbf{Rewrite Requirement:}\\
You may add, remove, replace, or reorganize any logic necessary to realize the target idea completely. Preserve parent mechanisms only when they remain consistent with the target idea. Do not introduce unrelated mechanisms.

\medskip
\textcolor{red}{Faithfully instantiate the decision principles, information signals, interactions, conditional behavior, and adaptive mechanisms described in the target idea. Do not introduce additional heuristic mechanisms that are not implied by the target idea.}

\medskip
When low-level implementation details are unspecified, choose the simplest implementation consistent with the target idea.

\medskip
\textbf{Output Requirement:}\\
Return only valid Python code wrapped in \texttt{```python ... ```}.

\end{rewritepromptbox}

\newpage
\subsection{Generated Heuristics}
\label{sec:generated heuristics}
This subsection provides example outputs of PRS-designed heuristics. The problem settings include a step-by-step constructive framework for the online bin packing problem and the ant colony optimization framework for the orienteering problem.

\definecolor{CodePurple}{RGB}{110,66,193}
\definecolor{CodePurpleDark}{RGB}{85,45,160}
\definecolor{CodeBg}{RGB}{247,247,250}
\definecolor{CodeKeywordBlue}{RGB}{30,102,245}
\definecolor{CodeTypeBlue}{RGB}{0,92,180}
\definecolor{CodeCommentGray}{RGB}{120,120,120}
\definecolor{CodeStringGreen}{RGB}{30,130,76}
\definecolor{CodeGreen}{RGB}{18,98,62}       
\definecolor{CodeGreenDark}{RGB}{16,92,56}    
\definecolor{CodeGreenDeepDark}{RGB}{6, 70, 40}
\lstdefinestyle{pyblue}{
  language=Python,
  basicstyle=\normalsize\ttfamily\footnotesize,
  columns=fullflexible,
  keepspaces=true,
  showstringspaces=false,
  breaklines=true,
  breakatwhitespace=true,
  tabsize=4,
  numbers=none,
  numberstyle=\scriptsize\color{CodeCommentGray},
  keywordstyle=\color{CodeKeywordBlue},     
  commentstyle=\color{CodeCommentGray}\bfseries,
  stringstyle=\color{CodeStringGreen},
  morekeywords=[2]{np,ndarray,float,int,bool,str,list,dict,tuple,set},
  keywordstyle=[2]\color{CodeTypeBlue},
}

\newtcblisting{codeexamplebox}[1][]{%
  enhanced,
  width=\linewidth,
  center,
  colback=CodeBg,
  colframe=CodeGreenDeepDark,
  colbacktitle=CodeGreenDeepDark,
  coltitle=white,
  fonttitle=\sffamily\bfseries\small,
  arc=6pt,
  boxrule=1.2pt,
  left=6pt,right=6pt,top=6pt,bottom=6pt,
  listing only,
  listing options={style=pyblue},
  title={Code Example},
  #1
}
\begin{codeexamplebox}[title={Evolved Heuristic \#1: priority function for OBPP with step-by-step constructive framework}]
def priority(item: float, bins_remain_cap: np.ndarray) -> np.ndarray:
    """
    This heuristic scores feasible bins using residual capacity, balanced utilization, and protection against premature bin exhaustion. It favors tight placements and moderate post-placement fill levels while giving additional preference to near-exact fits and already utilized bins.
    """
    import numpy as np
    eps = 1e-9
    tight_threshold = item * 0.1
    remaining = bins_remain_cap - item
    capacity_scale = bins_remain_cap.max() if bins_remain_cap.size > 0 else 1.0
    normalized_remaining = remaining / capacity_scale
    post_fill = 1 - normalized_remaining
    desired_fill = 0.5

    # Favor small residual capacity and balanced post-placement utilization
    residual_score = 1 / (remaining + eps)
    fill_gap = post_fill - desired_fill
    balance_score = np.exp(-20 * (fill_gap ** 2))

    # Discourage opening or repeatedly using nearly empty bins
    unused_penalty = np.exp(-10 * (normalized_remaining > (1 - item / capacity_scale)) * normalized_remaining)

    # Reward tight placement, near-exact fit, and previously utilized bins
    tight_bonus = np.where(
        remaining <= tight_threshold,
        1.5 + (tight_threshold - remaining) / tight_threshold,
        1.0
    )
    utilization = 1 - (bins_remain_cap / capacity_scale)
    utilization_bonus = 1 + 0.5 * utilization
    exact_match = (remaining < eps * 10).astype(float)
    exact_bonus = 1 + 0.2 * exact_match

    scores = residual_score * balance_score * unused_penalty * tight_bonus * utilization_bonus * exact_bonus
    return scores
\end{codeexamplebox}

\newtcblisting{codeexampleboxcont}[1][]{%
  enhanced,
  colback=CodeBg,
  colframe=CodeGreenDeepDark,
  colbacktitle=CodeGreenDeepDark,
  coltitle=white,
  fonttitle=\sffamily\bfseries\small,
  arc=6pt,
  boxrule=1.2pt,
  left=6pt,right=6pt,top=6pt,bottom=6pt,
  listing only,
  listing options={style=pyblue},
  title={Evolved Heuristic \#2 (cont.)},
  #1
}
\begin{codeexamplebox}[title={Evolved Heuristic \#2: Heuristic Matrix for OP with ACO framework}]
def heuristics(prize: "np.ndarray", distance: "np.ndarray", maxlen: float) -> "np.ndarray":
    """
    Edge-scoring heuristic that integrates reward, distance, and local neighborhood structure under a route-length constraint. It favors profitable, nearby, and well-connected transitions while suppressing moves that are difficult to complete feasibly.
    """
\end{codeexamplebox}

\begin{codeexampleboxcont}
    import numpy as np
    eps = 1e-9
    n_nodes = prize.shape[0]
    depot_dist = distance[:, 0]
    prize_scale = np.max(prize[1:]) if n_nodes > 1 else 1.0
    scaled_prize = prize / (prize_scale + eps)
    inverse_dist = np.zeros_like(distance)
    valid_dist = (distance > 0)
    inverse_dist[valid_dist] = 1.0 / (distance[valid_dist] + eps)
    
    prize_power, dist_power, density_power, connect_power = 7.0, 5.0, 4.0, 3.0
    depot_weight, reward_threshold = 2.2, 0.25
    local_radius = maxlen / 4.0
    high_reward = (scaled_prize >= reward_threshold)
    local_mask = (distance <= local_radius) & (distance > 0)
    local_mask = local_mask & high_reward[None, :]
    local_density = local_mask.sum(axis=1)
    density_scale = np.max(local_density) if np.max(local_density) > 0 else 1.0
    normalized_density = local_density / density_scale
    density_bonus = 1.0 + normalized_density  # reward nodes near many high-prize neighbors

    connect_radius = maxlen / 2.0
    connect_mask = (distance <= connect_radius) & (distance > 0) & high_reward[None, :]
    node_connectivity = connect_mask.sum(axis=1)
    connectivity_scale = np.max(node_connectivity) if np.max(node_connectivity) > 0 else 1.0
    normalized_connectivity = node_connectivity / connectivity_scale
    weak_node_penalty = np.where(normalized_connectivity < 0.3, 0.4, 1.0)
    connectivity_penalty = weak_node_penalty[:, None] * weak_node_penalty[None, :]  # downweight weakly connected nodes

    return_length = distance + depot_dist[None, :]
    normalized_depot_dist = depot_dist / (maxlen + eps)
    route_depth = (return_length + depot_dist[:, None]) / (maxlen + eps)
    route_depth = np.clip(1.0 - route_depth, 0.0, 1.0)
    infeasible_edges = return_length > maxlen  # forbid edges that cannot return to depot within maxlen

    depot_affinity = 1.0 - normalized_depot_dist
    depot_affinity = np.clip(depot_affinity, 0.0, 1.0)
    depot_bonus = 1.0 + depot_weight * ((depot_affinity[:, None] + depot_affinity[None, :]) / 2.0)

    edge_prize = (scaled_prize[:, None] + scaled_prize[None, :]) / 2.0
    ring_index = np.floor(normalized_depot_dist * 3).astype(int)
    ring_distance = np.abs(ring_index[:, None] - ring_index[None, :])
    ring_penalty = np.where(ring_distance > 1, 0.5, 1.0)

    structure_score = ((density_bonus[:, None] * density_bonus[None, :]) ** density_power * (connectivity_penalty ** connect_power))
    
    heuristic_matrix = ((edge_prize ** prize_power) * (inverse_dist ** dist_power) * structure_score * route_depth * depot_bonus * ring_penalty)
    heuristic_matrix[infeasible_edges] = 0.0
    np.fill_diagonal(heuristic_matrix, 0.0)
    heuristic_matrix = np.nan_to_num(heuristic_matrix, nan=eps, posinf=eps, neginf=eps)
    heuristic_matrix = np.maximum(heuristic_matrix, 0.0)
    return heuristic_matrix
\end{codeexampleboxcont}

\begin{codeexamplebox}[title={Evolved Heuristic \#3: Black-Box Heuristic (Edge Scoring) for OP with ACO framework}]
def heuristics(
    node_attr: np.ndarray, edge_attr: np.ndarray, node_constraint: float
) -> np.ndarray:
    """
    This heuristic ranks edges using endpoint attributes and inverse edge cost,
    with a soft penalty when the associated node-based constraint is exceeded.
    """
    n_nodes = len(node_attr)
    score_matrix = np.zeros((n_nodes, n_nodes))
    for i in range(n_nodes):
        for j in range(n_nodes):
            if edge_attr[i, j] > 0:
                combined_cost = np.sum(edge_attr[i]) + np.sum(edge_attr[:, j])
                base_score = (node_attr[i] + node_attr[j]) / edge_attr[i, j]
                if combined_cost > node_constraint:
                    score_matrix[i, j] = base_score * 0.5
                else:
                    score_matrix[i, j] = base_score
            else:
                score_matrix[i, j] = -1
    score_matrix = np.clip(score_matrix, -1, None)
    score_matrix = (score_matrix - np.mean(score_matrix)) / (np.std(score_matrix) + 1e-10)
    return score_matrix
\end{codeexamplebox}

\section{Details of Evaluation Configurations}
\label{sec:evaluation_details}

\textbf{Hyperparameters for PRS.} 
To initialize the heuristic pool, we set the number of initially sampled heuristics to 30, aligning with ReEvo \citep{ye2024reevo}. The softmax temperature is fixed to $T_{sel}=1.0$ for all tasks.
The base temperature of the LLMs for both perturbation and rewrite is 1.0. During initialization, the LLM temperature is further increased by 0.2, to enhance diversity. Our code will be released upon acceptance.

For adaptive operator control, we set the initial and final UCB exploration coefficients to $c_0=2.0$ and $c_T=0.5$, respectively, with a sliding reward window of 50 operator selections. The stagnation threshold and the post-restart operator reheating horizon are set to
$C=0.1T$ and $\Delta=0.1T$, respectively. We set an evaluation timeout of 60s for each heuristic in PRS following MCTS-AHD \citep{zheng2025monte}. For FSSP-GLS, this timeout is applied per instance within each heuristic evaluation.

\textbf{Use of Large Language Models.}
For most experiments, we use GPT-5-nano-2025-08-08 under its default reasoning mode to ensure consistency. 
In some problem settings, we additionally adopt GPT-4.1-nano to examine the adaptability of PRS. 
We further evaluate both closed-source and open-source LLMs to demonstrate the applicability and flexibility of our framework, with additional results reported in the Appendix \ref{sec:different_llms}. In designing heuristics under the ACO framework for CVRP, PRS uses on average 3.09M input tokens and 1.55M output tokens per run, with 1353 LLM queries and a total wall-clock time of 6h. Detailed runtime, token, and query-cost comparisons are reported in Appendix \ref{sec:cost_analysis}.


\textbf{Generation of Validation Sets $D$ and Testing Setups}. We follow the procedures in FunSearch \citep{romera2024mathematical}, EoH \citep{liu2024evolution}, ReEvo \citep{ye2024reevo} and MCTS-AHD to construct all validation datasets $D$ and testing instances. In particular, MCTS-AHD employs a varying-scale validation dataset $D$ for online BPP to facilitate the discovery of heuristics with improved generalization performance on a broader scale of testing instances, following the settings of ELG \citep{gao2023towards}. 
The complete evaluation configurations are summarized as follows:

\begin{itemize}
    \item \textbf{ACO Framework}. We adopt the configurations (i.e., number of ants and iterations) in ReEvo \citep{ye2024reevo}. \textbf{(a)} For CVRP (both black- and white-box settings), PRS is validated on a dataset of 10 instances with problem size 50 and tested on three sets of 64 instances each at size 50, 100, and 200. During both validation and testing, the number of ants and the number of iterations are fixed to 30 and 100, respectively. \textbf{(b)} For OP (both black- and white-box), the size of the dataset $D$ is 5 and the validation problem size is also 50. As for evolved heuristics for OP, they are tested with three sets of 64 instances each at size 50, 100, and 200. The number of ants is 20 while the number of iterations is 50 for both validation and testing. \textbf{(c)} For black-box MKP, the validation dataset consists of 5 instances of problem size 100. The testing set consists of two sets of 64 instances each at sizes 100 and 200. The number of ants is 10 and the number of iterations is 50 and 100 for validation and testing, respectively.
    \item \textbf{Step-by-step Construction Framework}.
    \textbf{(a)} For KP, PRS is validated on a dataset $D$ of 64 instances with problem size 100. Upon testing, the evolved heuristic is tested on three sets of 1,000 instances each at size 50, 100, 200. The capacity of size 50 is 12.5 while the capacity of size 100 and 200 is 25. \textbf{(b)} For Online BPP, PRS generates a dataset $D$ consisting of 4 instances: a 1,000-item instance with $W=100$, a 1,000-item instance with $W=500$, a 5,000-item instance with $W=100$, and a 5,000-item instance with $W=500$. The number of problem instances for each test set in Table \ref{tab:obpp} is 5. \textbf{(c)} For TSP, we use a validation set of 64 TSP instances with $N=50$ nodes and four test sets of 1,000 instances each at $N=50$, $100$, $200$, and $500$. \textbf{(d)} For ASP, PRS uses a validation set $D$ consisting of a single instance with $(n,w)=(15,10)$. Upon testing, the evolved heuristic is evaluated on four settings, namely $(12,7)$, $(15,10)$, $(21,15)$, and $(24,17)$, with one instance for each setting.
    \item \textbf{GLS Framework.} \textbf{(a)} For FSSP, we follow EoH to generate $D$: a dataset of 64 instances is randomly generated with a fixed number of 50 jobs and the number of machines is randomly chosen from $U[2,20]$. Details of the testing instances are provided in Appendix \ref{sec:full_fssp_results}. The maximum number of iterations is 1,000. The per instance timeout is 60s. \textbf{(b)} For TSP, we follow ReEvo to generate $D$ of 64 instances of problem size 200 and keep the same iteration limits and perturbation moves, which are 1200 and 30, respectively. We evaluated the heuristics on a wide range of instances as shown in Table \ref{tab:tsp_gls_ahd}. The details of the hyperparameters chosen for GLS are provided in Table \ref{tab:gls_params}.
\end{itemize}

\textbf{Hyperparameters for Baselines}. We use the default hyperparameters reported in the original papers for the baseline AHD methods whenever possible. To ensure a fair comparison, we standardize the maximum number of evaluated heuristics across all methods, using $T=2000$ for online BPP and $T=1000$ for all remaining tasks. For EoH, we retain its default population sizes of $20$ for online BPP and $10$ for TSP and FSSP, but do not impose the default generation limit; instead, the search continues until the common evaluation budget is reached.

\begin{table}[b]
\vspace{2mm}
\centering
\small
\setlength{\tabcolsep}{6pt}
\caption{GLS parameter settings for different TSP problem scales.
The number of perturbation moves controls the strength of solution diversification,
while the iteration limit determines the overall local search budget. Heuristics for TSP under GLS found by all LLM-based AHD frameworks are tested under the same configurations here.}
\label{tab:gls_params}
\begin{tabular}{c|cc}
\toprule
Problem Size & Perturbation Moves & Iteration Limit \\
\midrule
TSP20   & 5   & 73   \\
TSP50   & 40  & 300  \\
TSP100  & 40  & 1800 \\
TSP200  & 40  & 1000  \\
TSP500  & 60  & 250  \\
TSP1000 & 80  & 100   \\
\bottomrule
\end{tabular}
\vspace{-2mm}
\end{table}

\section{More Experimental Results}
\label{sec:extra_experiments}

\subsection{Search Dynamic Analysis}
\label{sec:search_dynamic_analysis}

\subsubsection{Diversity Analysis}
\label{sec:diversity_analysis}

Diversity in evolutionary search provides valuable insights into the exploration strength and the convergence performance of AHD algorithms \citep{pires2014diversity}. A substantial body of optimization research has investigated the impact of estimating and maintaining diversity within the evolutionary search \citep{pires2019dynamic,linlin2012diversity,dat2025hsevo}. The evolution curves alone in Section \ref{sec:discussion} are not sufficient to quantify the diversity within the heuristic search process of our PRS framework. Therefore, we adopt Cumulative Diversity Index (CDI) to quantify diversity, which is the same metric used in previous work, HSEvo \citep{dat2025hsevo}. In the context of heuristic search, CDI reflects how widely heuristic candidates are dispersed in the search space, serving as an implicit signal of system energy, or diversity \citep{jost2006entropy}.

\begin{figure*}[t]
  \centering
  \includegraphics[width=0.98\textwidth]{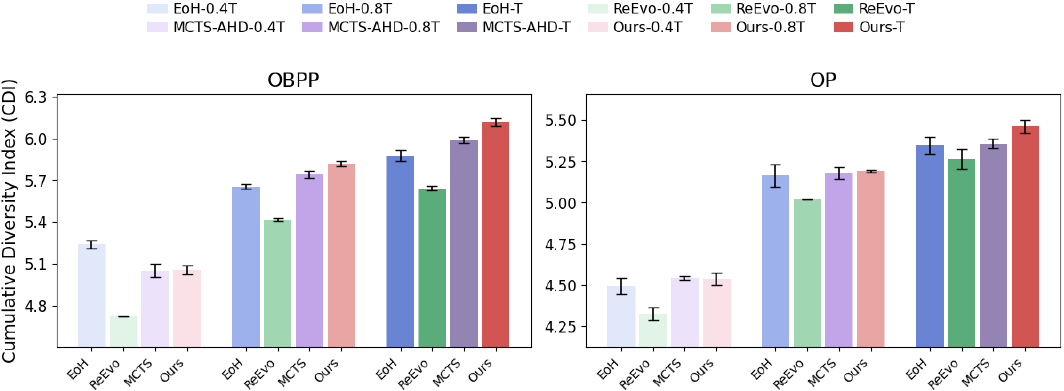}
  \caption{Diversity analysis (CDI) of EoH, ReEvo, MCTS-AHD, and PRS on OBPP-Constructive and OP-ACO at different search stages ($0.4T$, $0.8T$, and $T$). Error bars indicate standard deviation over three independent runs.}
  \label{fig:diversity_appendix}
\end{figure*}

To align with the approaches in calculating CDI in HSEvo, we utilized the cumulative archive of the embeddings of the top-20 candidates in every 80 evaluation steps to measure CDI for three other AHD baselines (i.e., EoH, ReEvo, and MCTS-AHD) and our PRS framework. CDI is calculated by constructing a Minimum Spanning Tree (MST) of all embeddings in the archive $A$ with edge lengths measured by Euclidean distance. The MST is then used to assess heuristic diversity. Let $d_i$ denote the length of the $i$-th edge in the minimum spanning tree (MST) constructed over the archive $A$, where $i \in \{1, 2, \ldots, |A|-1\}$. The cumulative diversity is computed using the Shannon entropy as
\begin{equation}
D_{\mathrm{CDI}} = - \sum_{i=1}^{|A|-1} p_i \log p_i .
\end{equation}
where the corresponding probability distribution of distance $d_i$ is defined as
\begin{equation}
p_i = \frac{d_i}{\sum_{j=1}^{|A|-1} d_j}, \quad i = 1, \ldots, |A|-1.
\end{equation}
A higher CDI reveals a more diverse dispersion of heuristic structures in the search space, and thus suggests more robust exploration dynamics. We mainly consider three stages to demonstrate the diversity of search in different phases of heuristic search: $\frac{2}{5}T$, $\frac{4}{5}T$, and $T$ within the maximum evaluation steps $T$, corresponding to the early, middle, and final stages of heuristic search ($T=1000$ for OP and $T=2000$ for OBPP).


As shown in Figure~\ref{fig:diversity_appendix}, all four AHD frameworks exhibit ascending CDI values as the evaluation steps grow, indicating that their cumulative heuristic archives become progressively more diverse during search. Across both OBPP and OP, PRS consistently achieves the highest or near-highest CDI at the middle and final stages and attains the largest final stage diversity among all compared methods. Compared to other baselines, PRS's diversity growth is particularly pronounced from $0.8T$ to $T$, suggesting that PRS maintains active exploration even in the late stage of the search rather than rapidly collapsing into a narrow set of heuristic structures. This delayed convergence indicates a stronger capability to avoid premature convergence and preserve diverse heuristic candidates throughout the search process.

\subsubsection{Idea-Space Diversity Analysis}
CDI measures diversity in the learned representation space of realized heuristic implementations, while mean nearest-neighbor distance (MNND) measures diversity in the semantic idea space before code realization. This distinction is important for PRS because its directional perturbation primitives operate directly on natural language heuristic ideas. Distance-based criteria have long been used to characterize dispersion and space-filling properties in multidimensional spaces \citep{johnson1990minimax,morris1995exploratory}, while mean nearest-neighbor distance provides a direct measure of local packing and redundancy \citep{layman2007stable}.  In the context of AHD, MNND characterizes how semantically distinct each retained heuristic idea is from its closest alternative in the archive: a larger value indicates fewer locally redundant ideas and broader coverage of distinct conceptual neighborhoods in the heuristic design space.
For each distinct heuristic idea $z_i$ in the cumulative archive, we obtain an L2-normalized sentence embedding $\mathbf{e}_i$ using \texttt{sentence-transformers/all-mpnet-base-v2}. The nearest-neighbor semantic distance of $z_i$ is defined using cosine distance, and MNND is computed as
\begin{equation}
D_{\mathrm{MNND}}
=
\frac{1}{N}
\sum_{i=1}^{N}
\min_{j \neq i}
\left(
1 -
\frac{\mathbf{e}_i^\top \mathbf{e}_j}
{\|\mathbf{e}_i\|_2 \|\mathbf{e}_j\|_2}
\right),
\end{equation}
where $N$ denotes the number of distinct ideas in the archive. By focusing on each idea's nearest semantic neighbor, MNND reflects local redundancy and sustained exploration across distinct regions of the heuristic idea space.

We follow the same archive construction protocol as in the CDI analysis, adding the top-20 valid candidates every 80 evaluations to a cumulative archive. Exact duplicate ideas are counted only once for MNND to avoid bias from repeated code realizations. The same treatment is applied to all compared methods. We evaluate PRS against EoH and MCTS-AHD, both of which explicitly maintain natural language heuristic descriptions. To minimize differences caused merely by the wording or style of the thought-generation prompts, we additionally standardize the natural language idea-generation prompt across the three methods while retaining their original search mechanisms. Thus, the comparison primarily reflects differences in how the respective search frameworks explore heuristic ideas rather than differences in description style. Results are averaged over three independent runs, with shaded regions denoting standard deviation.

\begin{figure*}[t]
    \centering
    \begin{subfigure}[t]{0.48\textwidth}
        \centering
        \includegraphics[width=\linewidth]{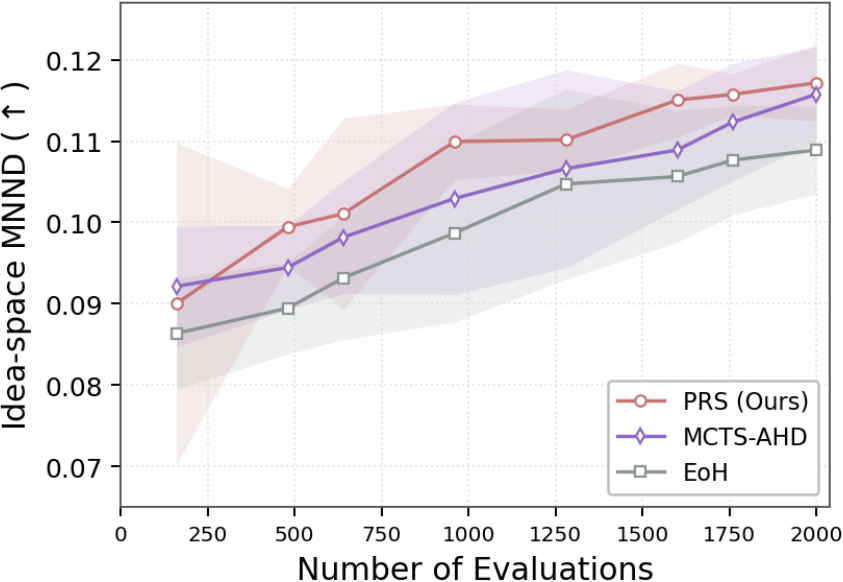}
        \caption{OBPP-Constructive}
    \end{subfigure}
    \hfill
    \begin{subfigure}[t]{0.48\textwidth}
        \centering
        \includegraphics[width=\linewidth]{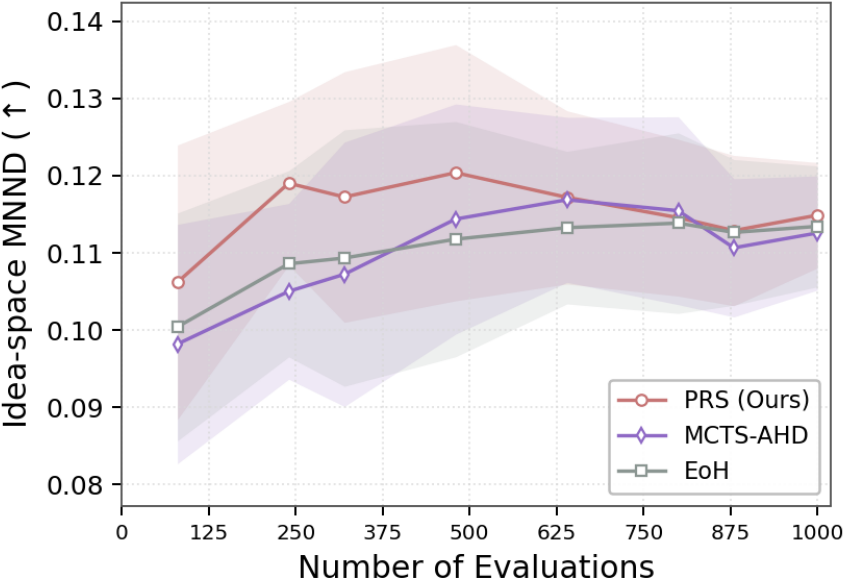}
        \caption{OP-ACO}
    \end{subfigure}
    \caption{Idea-space diversity measured by mean nearest-neighbor distance (MNND) for PRS, MCTS-AHD, and EoH. Higher values indicate that retained heuristic ideas remain farther from their closest semantic neighbors.}
    \label{fig:idea_mnnd}
\end{figure*}

As shown in Figure~\ref{fig:idea_mnnd}, PRS generally explores less locally redundant regions of the idea space than EoH and MCTS-AHD across both tasks. On OBPP-Constructive, although MCTS-AHD shows a slightly larger MNND at the earliest checkpoint, PRS subsequently maintains the highest mean MNND and finishes at approximately $0.117$, compared with $0.116$ for MCTS-AHD and $0.109$ for EoH. On OP-ACO, PRS shows a clearer advantage during the early and middle stages, reaching an MNND of approximately $0.120$, while EoH and MCTS-AHD remain lower; the gap narrows toward the end, but PRS still retains the highest final mean MNND. These results indicate that the directional perturbation primitives allow PRS to reach distinct semantic neighborhoods earlier and sustain broader idea space exploration throughout the search.

Together with the CDI results, the MNND analysis provides two complementary views of search diversity. CDI shows that PRS maintains broad dispersion among the realized heuristic implementations, whereas MNND shows that this diversity is already present at the level of semantic heuristic ideas. Their consistent trends suggest that the broader exploration of PRS is not merely a consequence of superficial variation in code realization. Instead, the explicit expansion, compression, and steering primitives promote sustained exploration of distinct regions of the heuristic design space before these ideas are instantiated as executable heuristics.

\subsubsection{Operator Contribution Analysis}
\label{sec:operator_contribution_analysis}

\begin{figure*}[t]
  \centering
  \includegraphics[width=1\textwidth]{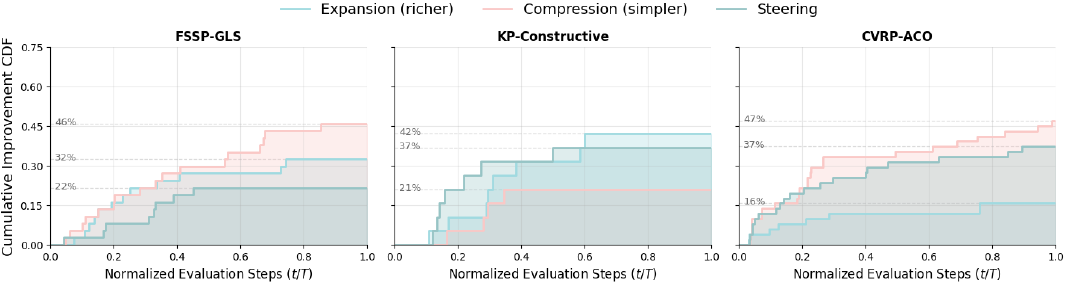}
  \caption{Operator contribution statistics. Each curve represents the cumulative share of best-so-far improvements triggered by each perturbation operator across all runs.}
  \label{fig:operator_count}
\end{figure*}

In this section, we further analyze how the three directional perturbation operators complement each other during search and how their contributions vary across different tasks and search stages. We measure the cumulative share of best-so-far improvements triggered by each operator across three representative settings: FSSP-GLS, KP-Constructive, and CVRP-ACO, based on five independent runs for each setting (Figure~\ref{fig:operator_count}).

1) \textit{FSSP-GLS.} In FSSP-GLS, compression is the dominant contributor, accounting for approximately 46\% of the cumulative improvements, and its contribution continues to increase throughout the search trajectory. Expansion contributes about 32\%, with most improvements occurring during the earlier and middle stages, while steering accounts for the remaining 22\%. This pattern suggests that both enriching and simplifying heuristic structures are useful in FSSP-GLS, with compression playing a particularly persistent role throughout the search.

2) \textit{KP-Constructive.} In KP-Constructive, the pattern changes substantially: expansion becomes the largest contributor at approximately 42\%, while steering also contributes strongly at around 37\%. Compression accounts for a smaller share of approximately 21\%, with most of its improvements occurring relatively early in the search. The comparable contributions of expansion and steering indicate that KP benefits not only from exploring richer heuristic structures, but also from fine-grained exploitation around promising regions.

3) \textit{CVRP-ACO.} In CVRP-ACO, compression again becomes the dominant contributor, accounting for approximately 47\% of cumulative improvements, followed by steering at around 37\%, whereas expansion contributes about 16\%. Compression accumulates improvements across the search trajectory, while steering also remains active across multiple stages. This result shows that effective search in CVRP-ACO relies more strongly on structural simplification and local refinement than on continued structural expansion.

\textbf{Stage-wise UCB behavior.}
To further examine the stage-wise behavior of adaptive operator selection, Table~\ref{tab:ucb_operator_behavior} reports the stage-wise behavior of the UCB controller. For each operator, we consider its selection rate, the probability of producing at least one improving beam child, and the mean best-child relative improvement. The results show that contribution is not determined by selection frequency alone. Across tasks and search stages, operators with similar selection rates can exhibit markedly different improvement probabilities and improvement magnitudes, indicating that the three perturbation directions provide distinct search utility even when they receive comparable amounts of search effort.

The stage-wise statistics also reveal clear task-dependent differences. In CVRP-ACO, compression maintains consistently strong improvement rates across later stages, which is consistent with its large cumulative contribution. In FSSP-GLS, compression frequently achieves high improvement rates, while steering often produces larger best-child improvements in several stages. For OBPP-Constructive, the relative utility of the operators changes more noticeably over time, with steering becoming particularly effective in the final stage. Together with Figure~\ref{fig:operator_count}, these results show that operator effectiveness varies across both tasks and search stages, supporting adaptive allocation among expansion, compression, and steering rather than a fixed operator preference.

\begin{table*}[t]
\centering
\small
\setlength{\tabcolsep}{2.5pt}
\renewcommand{\arraystretch}{1.08}
\caption{Stage-wise behavior of UCB-based operator selection. 
For each operator, we report the selection rate (Sel.), any-improvement rate (Imp.), 
and mean best-child relative improvement (Gain), all in \%. 
Statistics are aggregated over five runs.}
\label{tab:ucb_operator_behavior}

\resizebox{\textwidth}{!}{%
\begin{tabular}{llccc ccc ccc}
\toprule
\multirow{2}{*}{Task} &
\multirow{2}{*}{Evaluation stage} &
\multicolumn{3}{c}{Expansion} &
\multicolumn{3}{c}{Compression} &
\multicolumn{3}{c}{Steering} \\
\cmidrule(lr){3-5}
\cmidrule(lr){6-8}
\cmidrule(lr){9-11}
& & Sel. & Imp. & Gain
  & Sel. & Imp. & Gain
  & Sel. & Imp. & Gain \\
\midrule

\multirow{4}{*}{OBPP-Constructive}
& 0--249
& 32.42 & 35.21 & 3.241
& 34.25 & 26.67 & 5.881
& 33.33 & 41.10 & 4.849 \\
& 250--499
& 32.13 & 23.75 & 1.314
& 34.14 & 12.94 & 0.678
& 33.73 & 16.67 & 2.947 \\
& 500--749
& 34.15 & 20.24 & 3.541
& 32.11 & 16.46 & 2.415
& 33.74 & 22.89 & 1.958 \\
& 750--999
& 31.43 & 16.88 & 2.632
& 31.02 & 10.53 & 4.006
& \textbf{37.55} & \textbf{23.91} & \textbf{7.667} \\
\midrule

\multirow{4}{*}{CVRP-ACO}
& 0--249
& 31.36 & \textbf{66.67} & 3.947
& 34.09 & \textbf{66.67} & 4.059
& \textbf{34.55} & 53.95 & \textbf{6.751} \\
& 250--499
& 32.40 & 51.85 & \textbf{2.867}
& \textbf{34.40} & \textbf{70.93} & 2.171
& 33.20 & 45.78 & 2.513 \\
& 500--749
& 30.49 & 52.00 & 2.220
& \textbf{36.59} & \textbf{65.56} & \textbf{3.554}
& 32.93 & 43.21 & 3.106 \\
& 750--999
& 33.73 & 54.12 & 3.244
& \textbf{34.92} & \textbf{68.18} & \textbf{3.872}
& 31.35 & 60.76 & 3.269 \\
\midrule

\multirow{4}{*}{FSSP-GLS}
& 0--249
& 33.78 & 44.00 & 3.329
& 31.98 & \textbf{74.65} & 4.735
& 34.23 & 59.21 & \textbf{5.476} \\
& 250--499
& 31.98 & 56.96 & 3.557
& \textbf{34.01} & \textbf{61.90} & 3.236
& \textbf{34.01} & 51.19 & \textbf{4.500} \\
& 500--749
& 32.11 & 46.84 & 1.522
& \textbf{34.55} & 47.06 & 1.984
& 33.33 & \textbf{47.56} & \textbf{3.667} \\
& 750--999
& \textbf{35.34} & 44.32 & 3.309
& 31.33 & \textbf{64.10} & 1.388
& 33.33 & 46.99 & \textbf{3.499} \\
\bottomrule
\end{tabular}}
\end{table*}

\subsubsection{Operator-wise Generalization}
\label{sec:operator_generalization}

The preceding operator contribution analysis shows that expansion, compression, and steering contribute differently across tasks and search stages. We further examine whether the strongest candidates produced by different perturbation operators exhibit different in-distribution (ID) and out-of-distribution (OOD) generalization behavior.

For each independent run, we collect the valid children generated by each perturbation operator and rank them according to validation performance. Since AHD search ultimately prioritizes and retains high-performing candidates, we focus on the strongest outcomes produced by each perturbation direction and evaluate the top five children from each operator on both ID test instances and larger OOD instances.

\begin{table}[t]
\centering
\small
\caption{
Generalization performance of the top-five children generated by each perturbation operator.
Children are selected according to validation performance within each run.
Lower gap is better.
}
\label{tab:operator_generalization}
\setlength{\tabcolsep}{6pt}
\begin{tabular}{llcc}
\toprule
Task & Operator & ID gap $\downarrow$ & OOD gap $\downarrow$ \\
\midrule
KP-Constructive
& Expansion
& $\mathbf{0.03\%\pm0.01\%}$
& $0.04\%\pm0.02\%$ \\
& Compression
& $0.06\%\pm0.02\%$
& $\mathbf{0.03\%\pm0.00\%}$ \\
& Steering
& $0.05\%\pm0.02\%$
& $0.05\%\pm0.03\%$ \\
\midrule
CVRP-ACO
& Expansion
& $\mathbf{0.89\%\pm0.53\%}$
& $6.31\%\pm1.54\%$ \\
& Compression
& $1.45\%\pm0.71\%$
& $\mathbf{5.94\%\pm1.02\%}$ \\
& Steering
& $1.26\%\pm0.43\%$
& $6.52\%\pm0.63\%$ \\
\bottomrule
\end{tabular}
\end{table}

Table~\ref{tab:operator_generalization} shows a consistent qualitative difference between the strongest outcomes produced by the three perturbation directions. On KP-Constructive, expansion achieves the best ID performance, whereas compression achieves the best OOD performance. Steering remains competitive across both evaluation regimes.

A similar pattern appears on CVRP-ACO, where expansion again performs best on ID instances while compression achieves the strongest performance on larger OOD instances. Steering provides competitive intermediate performance without consistently favoring either regime.

These results do not imply that expansion is generally preferable for ID performance or that compression always improves OOD generalization. Rather, they suggest that different perturbation directions can produce strong heuristics under different evaluation conditions. 

\begin{table}[b]
\vspace{1mm}
\centering
\small
\setlength{\tabcolsep}{4pt}
\renewcommand{\arraystretch}{1.10}
\caption{
Performance of \textbf{ACO-based heuristics on black-box CVRP, MKP, and OP}.
We follow the settings of ReEvo \citep{ye2024reevo} in heuristic evolution and run each LLM-based method three times for average performance.
}
\label{tab:blackbox_single}

\begin{tabular}{l|cc|cc|cc}
\toprule
\multicolumn{1}{c|}{Task}
& \multicolumn{2}{c|}{CVRP}
& \multicolumn{2}{c|}{MKP}
& \multicolumn{2}{c}{OP} \\
\midrule
\multicolumn{1}{c|}{N=}
& \multicolumn{1}{c}{\underline{N=50}}
& \multicolumn{1}{c|}{N=100}
& \multicolumn{1}{c}{\underline{N=100}}
& \multicolumn{1}{c|}{N=200}
& \multicolumn{1}{c}{\underline{N=50}}
& \multicolumn{1}{c}{N=100} \\

\multicolumn{1}{c|}{Method}
& Obj.$\downarrow$ & Obj.$\downarrow$
& Obj.$\uparrow$ & Obj.$\uparrow$
& Obj.$\uparrow$ & Obj.$\uparrow$ \\
\midrule
ACO
& 18.282 & 30.030
& 18.969 & 38.766
& 13.570 & 24.149 \\
DeepACO
& 8.888 & 14.932
& 23.091 & 41.988
& 19.867 & 36.392 \\
\midrule
\multicolumn{7}{c}{\textit{LLM-based AHD (GPT-5-nano)}} \\
\midrule
EoH
& 9.360 & 16.700
& 22.518 & 40.568
& 14.776 & 28.627 \\
ReEvo
& 9.403 & 16.693
& 23.192 & 42.148
& 14.717 & 28.603 \\
MCTS-AHD
& 9.551 & 17.070
& 23.185 & 42.111
& \cellcolor{gray!20}\textbf{14.815} & 28.448 \\
Ours
& \cellcolor{gray!20}\textbf{9.351}
& \cellcolor{gray!20}\textbf{16.627}
& \cellcolor{gray!20}\textbf{23.207}
& \cellcolor{gray!20}\textbf{42.174}
& 14.789
& \cellcolor{gray!20}\textbf{28.898} \\
\bottomrule
\end{tabular}
\vspace{-2mm}
\end{table}

\subsection{Comparison on Black-box Settings}
\label{sec:appendix_blackbox}

We further evaluate PRS under black-box evaluation scenarios, which are inherently harder for heuristic search: candidate programs must remain executable under strict interfaces, and improvements must be discovered purely through outcome-driven trial-and-error, making robust diversification essential. Table~\ref{tab:blackbox_single} shows that PRS remains highly competitive or consistently exceeds other AHD frameworks on black-box tasks. These results suggest that PRS can sustain effective exploration even when supervision is extremely sparse, supporting robust heuristic search in opaque environments where objective evaluations are costly.

\subsection{TSP Step-by-step Constructive Framework}
\label{sec:TSP Step-by-step Constructive Framework}

For TSP, PRS is responsible for designing a function to select the next node to visit based on the node coordinates,
the starting point, the distance matrix, and all unvisited nodes.
The experiments in Table \ref{tab:tsp_step_by_step} test the heuristics on four different
scales, where PRS consistently achieves the best performance among the compared LLM-based AHD methods under both LLM settings.
For each method, we set a 1200s total time limit to ensure a manageable testing time budget and avoid excessive validation cost.

\begin{table}[t]
\vspace{1mm}
\centering
\small
\setlength{\tabcolsep}{2pt}
\renewcommand{\arraystretch}{1}
\caption{Performance of step-by-step construction heuristics on TSP, averaged over three runs. Bold denotes the best non-optimal objective value at each problem size. For each LLM setting, the best LLM-based AHD result in each column is shaded gray. Gaps are the relative difference to the optimal (or near-optimal) objective value obtained by LKH3.}
\label{tab:tsp_step_by_step}

\begin{tabular}{l|cc|cc|cc|cc}
\toprule
\multicolumn{9}{c}{\textbf{Travelling Salesman Problem (TSP)}}\\
\midrule
\multicolumn{1}{c|}{\textbf{Method}} &
\multicolumn{2}{c|}{\underline{TSP 50}} &
\multicolumn{2}{c|}{TSP 100} &
\multicolumn{2}{c|}{TSP 200} &
\multicolumn{2}{c}{TSP 500} \\
\multicolumn{1}{c|}{} &
Obj.$\downarrow$ & Gap$\downarrow$ &
Obj.$\downarrow$ & Gap$\downarrow$ &
Obj.$\downarrow$ & Gap$\downarrow$ &
Obj.$\downarrow$ & Gap$\downarrow$ \\
\midrule

LKH3~\citep{lin1973effective} & 5.680 & -- & 7.762 & -- & 10.715 & -- & 16.524 & -- \\
POMO &
\textbf{5.697} & \textbf{0.30\%} &
\textbf{8.001} & \textbf{3.08\%} &
12.897 & 20.37\% &
20.190 & 22.18\% \\
GC (Nearest Neighbour) &
6.944 & 22.26\% &
9.755 & 25.67\% &
13.502 & 26.02\% &
20.938 & 26.71\% \\
\midrule

\multicolumn{9}{c}{\textit{LLM-based AHD GPT-4.1-nano}}\\
\midrule
FunSearch &
6.583 & 15.90\% &
9.140 & 17.75\% &
12.798 & 19.44\% &
19.985 & 20.94\% \\
EoH &
6.345 & 11.71\% &
8.862 & 14.17\% &
12.428 & 15.99\% &
19.568 & 18.42\% \\
MCTS-AHD &
6.337 & 11.57\% &
8.865 & 14.21\% &
12.419 & 15.91\% &
19.382 & 17.29\% \\
PRS (Ours) &
\cellcolor{gray!20}6.299 & \cellcolor{gray!20}10.90\% &
\cellcolor{gray!20}8.797 & \cellcolor{gray!20}13.33\% &
\cellcolor{gray!20}12.268 & \cellcolor{gray!20}14.50\% &
\cellcolor{gray!20}19.092 & \cellcolor{gray!20}15.54\% \\
\midrule

\multicolumn{9}{c}{\textit{LLM-based AHD GPT-5-nano}}\\
\midrule
FunSearch &
6.444 & 13.46\% &
9.104 & 17.28\% &
12.874 & 20.15\% &
20.021 & 21.16\% \\
EoH &
6.377 & 12.27\% &
8.755 & 12.80\% &
12.563 & 17.25\% &
19.874 & 20.27\% \\
MCTS-AHD &
6.299 & 10.89\% &
8.694 & 12.00\% &
12.364 & 15.39\% &
18.825 & 13.92\% \\
PRS (Ours) &
\cellcolor{gray!20}6.215 & \cellcolor{gray!20}9.42\% &
\cellcolor{gray!20}8.568 & \cellcolor{gray!20}10.38\% &
\cellcolor{gray!20}\textbf{12.161} & \cellcolor{gray!20}\textbf{13.50\%} &
\cellcolor{gray!20}\textbf{18.553} & \cellcolor{gray!20}\textbf{12.28\%} \\
\bottomrule
\end{tabular}
\vspace{-1mm}
\end{table}

\textbf{Discussion.}
Across both LLM settings, PRS consistently outperforms all compared LLM-based AHD methods on TSP50, TSP100, TSP200, and TSP500.
In particular, PRS improves upon MCTS-AHD across all four problem sizes under both GPT-4.1-nano and GPT-5-nano, while the advantage remains evident as the problem scale increases.
The results also show that PRS benefits consistently from the stronger GPT-5-nano backbone, achieving lower gaps across all four test sizes.
Overall, these results demonstrate that the proposed multi-directional search process remains effective across different LLM backbones and generalizes well from the search setting to substantially larger TSP instances.

\subsection{TSP Heuristic under Guided Local Search Framework}
\label{sec:TSP Heuristic under Guided Local Search Framework}

To evaluate the effectiveness of PRS in designing penalty heuristics
within the guided local search (GLS) framework, we follow the general experimental
protocol of Ye et al.~\citep{ye2024reevo}.
Specifically, \textsc{PRS} is employed to generate the key penalty heuristic
used to construct the knowledge matrices in KGLS~\citep{arnold2019knowledge},
which enables adaptive refinement of the original manually designed heuristic.

All heuristic generation and evaluation procedures are conducted using GPT-5-nano as the backbone LLM.
We evaluate the resulting heuristics on different
problem scales of TSP, and compare against representative LLM-based AHD baselines, the original KGLS with handcrafted penalty functions, as well as two
neural combinatorial optimization (NCO) methods, namely NeuralOpt \citep{ma2021learning} and GNNGLS \citep{hudson2021graph}. NeuralOpt is an NCO method that learns to improve the solutions, while GNNGLS is primarily based on GLS framework. The GLS hyperparameters are scaled with problem size to balance exploration and computational cost.
Specifically, the number of perturbation moves and the iteration limit are adaptive to problem sizes, as summarized in Table~\ref{tab:gls_params}.
This adaptive budgeting ensures fair comparison across problem scales while avoiding excessive local search on large instances.

As shown in Table~\ref{tab:tsp_gls_ahd}, \textsc{PRS} consistently improves
the performance of KGLS across different problem scales.
In particular, \textsc{PRS} achieves lower optimality gaps on TSP50, TSP100 and TSP500
compared to other LLM-based AHD methods while maintaining competitive or the lowest runtime. Notably, our approach consistently reaches zero gaps on small-scale problems (TSP50 and TSP100) and achieves the lowest gap $1.649\%$ on TSP 500 and the second lowest on TSP 200.
Together with Table \ref{tab:tsp_step_by_step}, these findings further demonstrate that our multi-directional perturbation--rewrite framework is effective and adaptable across different heuristic formulations and problem scales.
\begin{figure*}[t]
    \centering
    \includegraphics[width=0.88\textwidth]{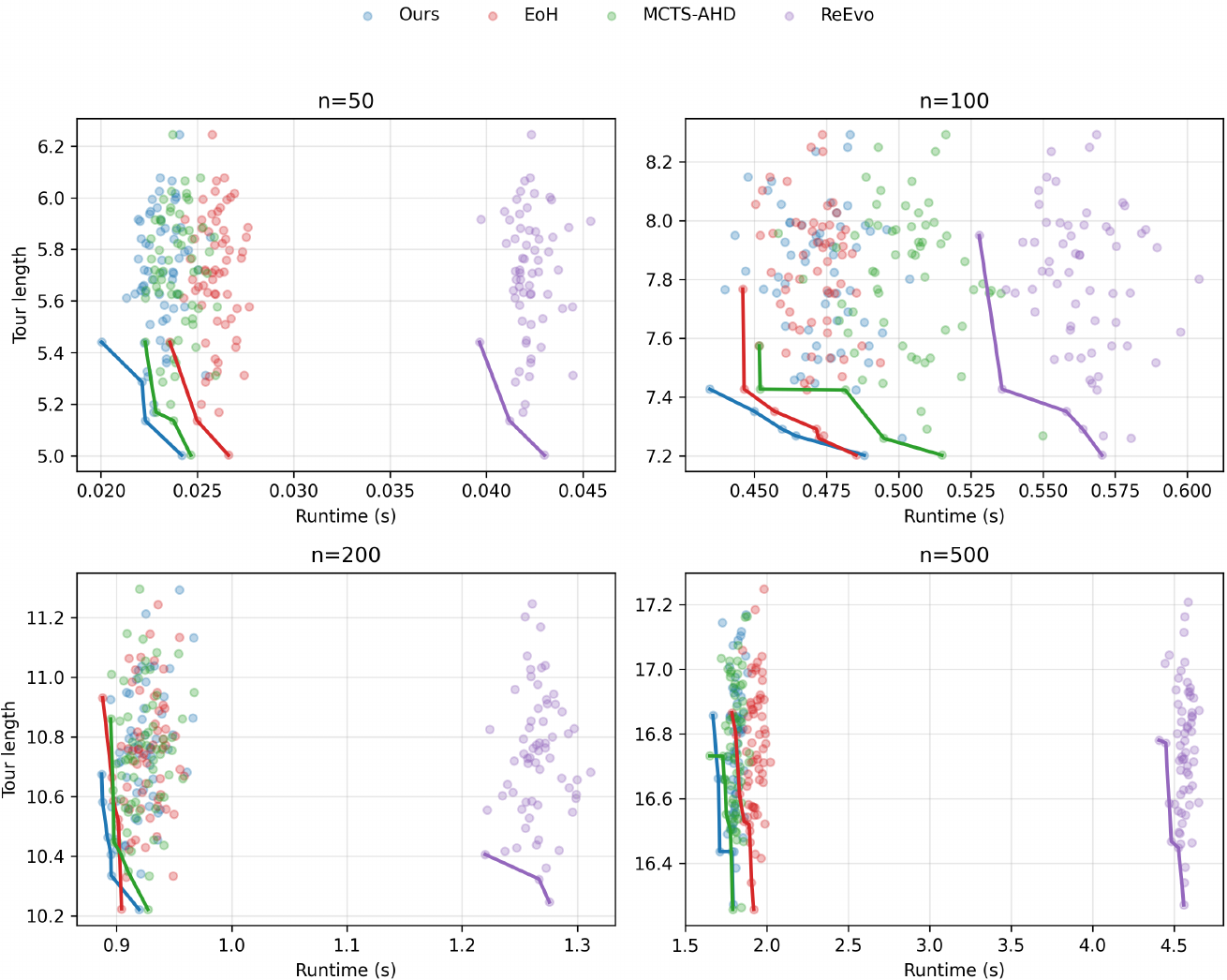}
    \caption{Pareto analysis of runtime and solution quality on TSP under the GLS framework.
    Each scatter point corresponds to one heuristic candidate evaluated during search, where the x-axis denotes runtime per instance and the y-axis denotes tour length.
    Results are shown for four problem sizes ($n=50,100,200,500$).
    The connected frontier for each method highlights the best observed quality--efficiency trade-off among its discovered heuristics.
    Lower tour length and runtime are preferred.}
    \label{fig:tsp_gls_pareto}
\end{figure*}
\begin{table*}[t]
\centering
\small
\setlength{\tabcolsep}{4pt}
\caption{Performance of heuristics on TSP benchmarks with guided local search framework. 
For a fair comparison, we report the result of the best performing heuristic over three independent runs following ReEvo \citep{ye2024reevo}. 
Gap (\%) denotes the relative optimality gap to the best-known solution, and Time (s) denotes the average runtime per instance. \textbf{*} denotes results drawn from ReEvo \citep{ye2024reevo}.
}
\label{tab:tsp_gls_ahd}
\begin{tabular}{l|cc|cc|cc|cc}
\toprule
Method
& \multicolumn{2}{c|}{TSP50}
& \multicolumn{2}{c|}{TSP100}
& \multicolumn{2}{c|}{\underline{TSP200}}
& \multicolumn{2}{c}{TSP500} \\
\cmidrule(lr){2-3} 
\cmidrule(lr){4-5} 
\cmidrule(lr){6-7}
\cmidrule(lr){8-9}
& Gap(\%)$\downarrow$ & Time(s)
& Gap(\%)$\downarrow$ & Time(s)
& Gap(\%)$\downarrow$ & Time(s)
& Gap(\%)$\downarrow$ & Time(s) \\
\midrule
NeuOpt*   
& 0.000 & 1.32 
& 0.027 & 2.67 
& 0.403 & 4.81 
& -- & -- \\

GNNGLS*   
& 0.052 & 3.83 
& 0.705 & 6.78 
& 3.522 & 9.92 
& -- & -- \\

\midrule
EoH\
& \textbf{0.000} & 0.03
& 0.390 & 0.47
& \cellcolor{gray!20}\textbf{0.196} & 0.90
& \underline{1.667} & 1.93 \\
ReEvo\
& \textbf{0.000} & 0.04
& 0.025 & 0.55
& 0.338 & 1.26
& 1.703 & 4.56 \\
MCTS-AHD\
& \textbf{0.000} & 0.02
& \textbf{0.000} & 0.47
& 0.270 & 0.90
& 1.680 & 1.82 \\
Ours\
& \cellcolor{gray!20}\textbf{0.000} & 0.02
& \cellcolor{gray!20}\textbf{0.000} & 0.46
& \underline{0.212} & 0.90
& \cellcolor{gray!20}\textbf{1.649} & 1.81 \\

\bottomrule
\end{tabular}
\end{table*}

\paragraph{Pareto analysis of runtime and solution quality.}
Beyond reporting the final best-performing heuristic in Table~\ref{tab:tsp_gls_ahd}, we further analyze the quality--efficiency tradeoff of the heuristics discovered by different AHD frameworks on TSP under the GLS framework.
Figure~\ref{fig:tsp_gls_pareto} visualizes the runtime and tour length of candidate heuristics discovered during search at four problem scales.

Across all scales, PRS shows a consistently favorable Pareto frontier.
At smaller scales ($n=50$ and $n=100$), the heuristics discovered by PRS achieve lower tour lengths while remaining among the fastest candidates.
At larger scales ($n=200$ and $n=500$), this advantage becomes more evident: PRS maintains a frontier that is competitive or better in solution quality while avoiding the large runtime overhead observed in some baselines, especially ReEvo.
This indicates that PRS is able to discover heuristics with a stronger balance between effectiveness and computational efficiency, rather than improving quality merely through more expensive heuristic structures.

The results also provide additional evidence for the discussion in Appendix \ref{sec:TSP Step-by-step Constructive Framework}.
PRS tends to discover candidates that remain closer to the efficient part of the quality--runtime tradeoff frontier, especially on medium- and large-scale TSP instances.

\begin{table*}[b]
\centering
\small
\setlength{\tabcolsep}{5pt}
\caption{Results on Taillard \citep{taillard1993benchmarks} instance sets. The results of non-LLM-based methods are cited from EoH \citep{liu2024evolution}. Each entry reports the average relative gap (\%) to the known upper bound in Taillard (lower is better). The best result in each row is highlighted in \textbf{bold}. Additionally, among all the LLM-based AHD methods, the best value is shaded.}
\label{tab:full_taillard_tab}
\begin{tabular}{lcccccccccc}
\toprule
Test Set & GUPTA & CDS & NEH & NEHFF & PFSPNet & LS & ILS1 & ILS2 & EoH & Ours \\
\midrule
j50\_m5   & 12.23 & 6.98 & 0.57 & 0.40 & 0.51 & 0.32 & 0.15 & 0.32 & \cellcolor{gray!20}\textbf{0.00} & \cellcolor{gray!20}\textbf{0.00} \\
j50\_m10  & 20.11 & 12.72 & 3.47 & 3.62 & 3.48 & 3.33 & 1.47 & 0.29 & 0.21 & \cellcolor{gray!20}\textbf{0.17} \\
j50\_m20  & 22.78 & 15.03 & 5.48 & 5.10 & 5.05 & 4.67 & 2.13 & \textbf{0.34} & 0.72 & \cellcolor{gray!20}0.83 \\
\hline
j100\_m5  & 5.98  & 5.10  & 0.39 & 0.31 & 0.31 & 0.28 & 0.20 & 0.38 & -0.04 & \cellcolor{gray!20}\textbf{-0.05} \\
j100\_m10 & 15.03 & 9.36  & 2.07 & 1.88 & 1.72 & 1.38 & 0.77 & 0.34 & 0.22 & \cellcolor{gray!20}\textbf{0.14} \\
j100\_m20 & 21.00 & 13.55 & 3.58 & 3.73 & 3.56 & 3.51 & 2.27 & 0.43 & 0.62 & \cellcolor{gray!20}\textbf{0.27} \\
\hline
j200\_m10 & 11.59 & 7.22  & 0.98 & 0.70 & 0.82 & 0.87 & 0.74 & 0.54 & 0.18 & \cellcolor{gray!20}\textbf{0.18} \\
j200\_m20 & 18.09 & 11.89 & 2.90 & 2.52 & 2.49 & 2.53 & 2.26 & 0.59 & 0.65 & \cellcolor{gray!20}\textbf{0.55} \\
\midrule
Average  & 15.85 & 10.23 & 2.43 & 2.28 & 2.24 & 2.11 & 1.25 & 0.40 & 0.32 & \cellcolor{gray!20}\textbf{0.26} \\
\bottomrule
\end{tabular}
\end{table*}

\subsection{Flow Shop Scheduling Problem under Guided Local Search Framework}
\label{sec:full_fssp_results}
We adopt the same guided local search (GLS) framework equipped with two neighborhood operators,
namely \emph{Swap} and \emph{Relocate} as EoH.
Within this framework, PRS is employed to automatically design a heuristic strategy
that jointly governs the update of the execution-time matrix and the selection of jobs to be perturbed during the search process.
The heuristics generated by PRS are compared against a diverse set of non-AHD methods, including both classical heuristics and learning-based approaches.
Specifically, we consider the following competitors:
(i) \emph{GUPTA}~\citep{gupta1971functional}, a classical heuristic for the flow-shop scheduling problem (FSSP);
(ii) \emph{CDS}~\citep{campbell1970heuristic}, another well-established heuristic for FSSP;
(iii) \emph{NEH}~\citep{nawaz1983heuristic}, which is widely regarded as an effective heuristic for flow-shop scheduling;
(iv) \emph{NEHFF}~\citep{fernandez2014insertion}, an enhanced variant of NEH;
(v) \emph{PFSPNet}~\citep{pan2021deep}, a recently proposed deep learning-based solver for flow-shop scheduling;
(vi) \emph{Local Search (LS)}, which applies the same neighborhood operators as PRS without heuristic adaptation;
(vii) \emph{ILS1}~\citep{stutzle1998applying}, a classical iterated local search method for FSSP; and
(viii) \emph{ILS2}, which follows the same overall framework as our method but relies on manually designed heuristic rules. We take EoH as the LLM-based AHD baseline. All methods are evaluated on the widely used Taillard \citep{taillard1993benchmarks} benchmark instances.
The number of jobs ranges from 50 to 200, while the number of machines varies between 5 and 20. Each reported value is the average relative gap over 10 instances in the corresponding test set.

As shown in Table \ref{tab:full_taillard_tab}, PRS achieves lower gaps across Taillard instances than EoH in most test sets (e.g., $0.27\%$ vs $0.62\%$ on j100\_m20), outperforming it on average while also demonstrating superior performance against non-AHD baselines, including NEH-based heuristics and iterated local search methods.

\subsection{Results on TSPLib instances}
\label{sec:tsplib_results}

We conduct experiments on the well-known real-world TSP benchmark TSPLib~\citep{reinelt1991tsplib}
to evaluate the effectiveness of the step-by-step constructive heuristics of PRS in generalizing to real-world TSP instances. 
We consider TSPLib instances with up to 500 cities following MCTS-AHD \citep{zheng2025monte} and report the optimality gap with respect to the best-known solutions.

We compare \textsc{PRS} against manually designed heuristics, including Greedy,
as well as the GP-based AHD method GHPP~\citep{duflo2019gp}.
For LLM-based AHD baselines, we include the best-performing heuristics of EoH and ReEvo.
All LLM-designed heuristics are executed three times with different starting nodes
and averaged for fair comparison. 
The results of the instances reported in GHPP within the problem scale $[0,500]$ are reported in Table \ref{tab:tsplib_step_gap}.
As shown in Table~\ref{tab:tsplib_step_gap}, the heuristic generated by \textsc{PRS} achieves a lower overall optimality gap than manually designed heuristics and the
GP-based AHD baseline, and leads all methods in 8 of the 15 instances. Collectively, these results  demonstrate the competitive performance of PRS for designing TSP-Constructive heuristics that generalize well to an out-of-distribution benchmark.

\begin{table}[H]
\vspace{-2mm}
\small
\centering
\setlength{\tabcolsep}{2pt}
\renewcommand{\arraystretch}{0.9}
\caption{Results of the GP-based method GHPP, and LLM-based AHD methods on designing step-by-step constructive
heuristics for TSP on TSPLib instances.
We report the optimality gap (\%) of each instance.
The results of GHPP-best, Christofides \citep{christofides2022worst}, Greedy \citep{brecklinghaus2015approximation}, Nearest insertion and Nearest-greedy \citep{rosenkrantz1977analysis}  are taken from~\citep{duflo2019gp}.
LLM-based AHD methods (EoH, ReEvo, and \textsc{PRS}) are each run three times with different
starting nodes and averaged.
The leading LLM-designed heuristic on each instance is shaded, and the overall best
heuristic is shown in bold.
}
\label{tab:tsplib_step_gap}
\begin{tabular}{l|cccc|cccc}
\noalign{\hrule height 1.2pt}
Instance & Christofides & Greedy & Nearest insertion &Nearest-greedy & GHPP-best & EoH & ReEvo &  Ours \\
\midrule
bier127  & 13.03 & 19.50 & 23.05& 23.25 & 15.64 & 19.60 & 10.79 & \cellcolor{gray!20}\textbf{8.31}  \\
d493     & \textbf{9.52} & 16.68 & 20.39& 24.00& 15.58 & 17.02 & 13.43 & \cellcolor{gray!20}13.15 \\
ch130    &10.09& 28.40 &19.21&25.66& 14.77 & 12.47 & \cellcolor{gray!20}\textbf{9.40} &12.24 \\
eil51    &15.18& 13.03 &16.14&31.96& 10.20 & 18.51 & 6.47  &  15.20  \\
kroA150  &13.44& 20.24 &19.09&26.08& 15.59 & 24.61 & 11.62 & \cellcolor{gray!20}\textbf{9.86}  \\
kroB100  &\textbf{9.82}& 16.59 &21.53&26.26& 14.06 & 13.46 & 12.20 &  \cellcolor{gray!20}10.56 \\
kroC100  & 9.08 & 12.94 &24.25&25.76& 16.22 & 16.85 & 15.88  & \cellcolor{gray!20}\textbf{8.16} \\
lin318   &13.80& 18.75 &24.44&25.78& 14.30 & 23.02 & 16.63 & \cellcolor{gray!20}\textbf{10.95} \\
pr226    &14.17& 21.44 &28.02&24.65& 15.50 & 12.44 & 18.02  & \cellcolor{gray!20}\textbf{10.00} \\
pr264    &11.28 & 11.89 &34.28&17.87&  24.00 & 21.50 & 16.78 & \cellcolor{gray!20}\textbf{10.97} \\
pr299    &\textbf{11.23} & 31.42 &25.05&31.42& 18.24 & 20.95 & 20.63 & \cellcolor{gray!20}15.06 \\
rat99    & 9.43 & 22.30 &21.05&21.79& 14.09 & 21.01 & 12.41 & \cellcolor{gray!20}\textbf{9.18}\\
ts225    &5.67 &\textbf{5.38} &19.93&16.82& 7.71  & 17.55 & \cellcolor{gray!20}6.56  & 11.33  \\
fl417     &15.57& 12.64 &25.52&32.42& 22.72 & 24.13 & 19.15 & \cellcolor{gray!20}\textbf{10.59} \\
pr439     &\textbf{11.16} & 20.08 &24.67&27.36& 21.36 & 26.83 & 19.25 & \cellcolor{gray!20}14.52 \\
\midrule
\noalign{\hrule height 1.2pt}
\end{tabular}
\end{table}

\subsection{Statistical Significance}
\label{sec:statistical_significance}

In this section, we provide a more rigorous statistical significance analysis on three representative scenarios: CVRP under the ACO framework, KP under the step-by-step constructive framework, and TSP with GLS at $N=200$.
We conduct 10 independent runs for CVRP-ACO and KP-Constructive, and 5 independent runs for TSP-GLS.
For each setting, we report the best objective score for each run, together with their average and standard deviation.
We then perform one-sided paired $t$-tests to compare PRS against the state-of-the-art baseline MCTS-AHD.
The results in Table \ref{tab:significance_test} indicate that the observed gains of PRS are statistically significant over MCTS-AHD across all three settings.
Moreover, the standard deviations of PRS remain comparable to those of MCTS-AHD, suggesting that its improvements are not obtained at the cost of substantially increased instability.

\begin{table}[H]
\centering
\footnotesize
\setlength{\tabcolsep}{0.3pt}
\renewcommand{\arraystretch}{1.05}
\caption{Statistical significance analysis of PRS over multiple independent runs. 
The reported p-values are from one-sided paired $t$-tests comparing PRS against MCTS-AHD. For TSP-GLS, the reported objective values denote gap (\%).}
\label{tab:significance_test}
\begin{tabular}{lccccccccccccc}
\thickhline
\midrule
Method & run1 & run2 & run3 & run4 & run5 & run6 & run7 & run8 & run9 & run10 & avg & std & p-value \\
\midrule

\multicolumn{14}{c}{\textbf{CVRP-ACO} ($\downarrow$)} \\
\midrule
MCTS-AHD & 8.842 & 8.957 & 8.623 & 8.993 & 9.012 & 8.790 & 9.103 & 8.929 & 9.231 & 9.186 & 8.967 & 0.184 & -- \\
PRS & 8.563 & 8.588 & 8.751 & 8.785 & 8.989 & 8.949 & 8.835 & 8.943 & 9.012 & 9.058 & \textbf{8.847} & 0.174 & 0.0331 \\

\midrule
\multicolumn{14}{c}{\textbf{KP-Constructive} ($\uparrow$)} \\
\midrule
MCTS-AHD & 40.249 & 40.255 & 40.261 & 40.257 & 40.258 & 40.252 & 40.258 & 40.254 & 40.260 & 40.252 & 40.256 & 0.004 & -- \\
PRS & 40.262 & 40.258 & 40.263 & 40.258 & 40.266 & 40.259 & 40.262 & 40.258 & 40.263 & 40.261 & \textbf{40.261} & 0.003 & 0.001 \\

\midrule
\multicolumn{14}{c}{\textbf{TSP-GLS} ($\downarrow$)} \\
\midrule
MCTS-AHD & 0.185 & 0.216 & 0.238 & 0.208 & 0.198 & -- & -- & -- & -- & -- & 0.209 & 0.020 & -- \\
PRS & 0.169 & 0.191 & 0.187 & 0.153 & 0.209 & -- & -- & -- & -- & -- & \textbf{0.182} & 0.021 & 0.0439 \\
\midrule
\thickhline
\end{tabular}
\end{table}

\subsection{Search Strategy Variants}
\label{sec:search_strategy_variants}

In this section, we examine whether the effectiveness of PRS is critically dependent on the use of BFS.
To this end, we replace the BFS-based search in PRS with two alternative strategies, namely CoT-SC~\citep{wang2022self} and MCTS, while keeping the perturbation--rewrite operator unchanged.
Table~\ref{tab:search_strategy_variants} shows that PRS-BFS achieves the best overall performance on both KP and CVRP.
However, the margins are moderate, and both PRS-MCTS and PRS-CoT-SC remain competitive.
This suggests that BFS is a strong default choice in our framework, likely because it provides a favorable balance between exploration coverage, candidate utilization, and search efficiency, but the effectiveness of PRS does not rely solely on BFS.
\begin{table}[h]
\centering
\small
\setlength{\tabcolsep}{6pt}
\renewcommand{\arraystretch}{1.08}
\caption{Comparison of search strategy variants within the PRS framework. Results are reported as mean $\pm$ standard deviation over three independent runs. For KP, higher is better; for CVRP, lower is better. The maximum number of evaluated heuristics is fixed to $T=1000$ for all variants.}
\label{tab:search_strategy_variants}
\begin{tabular}{lcc}
\toprule
\textbf{Variations} & \textbf{KP ($n=100$) $\uparrow$} & \textbf{CVRP ($n=50$) $\downarrow$} \\
\midrule
PRS-BFS
& \bmstd{40.2618}{0.0145}
& \bmstd{8.9583}{0.0726} \\

PRS-CoT-SC
& \mstd{40.2532}{0.0418}
& \mstd{9.3159}{0.0586} \\

PRS-MCTS
& \mstd{40.2584}{0.0175}
& \mstd{9.0728}{0.0960} \\
\bottomrule
\end{tabular}
\vspace{-2mm}
\end{table}

\newpage

\subsection{Experiments with different LLMs}
\label{sec:different_llms}
Tables \ref{tab:llms_obpp}-\ref{tab:llms_kp} further break down the performance of heuristics generated by different LLM backbones on OBPP and KP. We benchmark PRS on LLM models other than GPT, including a closed-source model Claude-3.7-sonnet and the open-source models, namely DeepSeek-v3 and Llama-3.1-70B-Instruct.

GPT-5-nano achieves the best performance among closed-source models on 1k\_100 and exhibits the smallest variance across both settings, while GPT-4.1-nano obtains a slightly lower gap on 5k\_500. Among open-source models, DeepSeek-v3 performs competitively and achieves the best result on the OOD larger-scale setting (5k\_500), while Llama-3.1-70B-Instruct exhibits slightly larger gaps. Overall, closed-source models tend to produce more stable heuristics on in-the-distribution instances, whereas strong open-source models can still achieve competitive performance on the OOD setting.

\begin{table}[h]
\vspace{-2mm}
\centering
\small
\setlength{\tabcolsep}{8pt}
\renewcommand{\arraystretch}{1.1}
\caption{Benchmarking PRS on different LLMs on OBPP (average gap (\%) to the lower bound). Best result among all methods of each problem size is highlighted in bold while the best result among all LLM-based results of each scale is shaded.}
\label{tab:llms_obpp}

\begin{tabular}{l|cc|cc}
\noalign{\hrule height 1.2pt}
\multicolumn{5}{c}{\textbf{OBPP}} \\
\noalign{\hrule height 0.6pt}
\multicolumn{1}{c|}{N=} & \multicolumn{2}{c|}{\underline{1k\_100}} & \multicolumn{2}{c}{5k\_500} \\
\cline{2-5}
\multicolumn{1}{c|}{Method} & Gap (\%) $\downarrow$ & std (\%) & Gap (\%) $\downarrow$ & std (\%) \\
\noalign{\hrule height 0.4pt}
Best Fit  & 4.77\% & -- & 0.55\% & -- \\
First Fit & 5.02\% & -- & 0.55\% & -- \\
\noalign{\hrule height 0.4pt}
\multicolumn{5}{c}{\textit{Closed Source LLMs}} \\
\noalign{\hrule height 0.4pt}
GPT-4.1-nano       & 2.58\% & 0.723 & 0.32\% & 0.125 \\
GPT-5-nano         & \cellcolor{gray!20}\textbf{1.82\%} & \cellcolor{gray!20}\textbf{0.123} & 0.35\% & 0.117 \\
Claude-3.7-sonnet  & 2.25\% & 0.489 & 0.39\% & 0.178 \\
\noalign{\hrule height 0.4pt}
\multicolumn{5}{c}{\textit{Open Source LLMs}} \\
\noalign{\hrule height 0.4pt}
DeepSeek-v3            & 3.40\% & 0.456 & \cellcolor{gray!20}\textbf{0.23\%} & \cellcolor{gray!20}\textbf{0.164} \\
Llama-3.1-70B-Instruct & 4.03\% & 0.361 & 0.40\% & 0.124 \\
\noalign{\hrule height 1.2pt}
\end{tabular}
\end{table}

On KP, all evaluated LLMs are capable of generating high-performing heuristics, which substantially outperform the constructive neural baseline POMO on both $N=100$ and $N=200$. In particular, GPT-5-nano and Claude-3.7-sonnet consistently achieve the smallest optimality gaps across both problem sizes, while open-source models such as DeepSeek-v3 and Llama-3.1-70B-Instruct remain slightly behind. These results suggest that the proposed PRS is robust to the choice of LLM backbones, while stronger models primarily contribute to improved stability and marginal gains in solution quality.

\begin{table}[h]
\vspace{-2mm}
\small
\centering
\setlength{\tabcolsep}{8pt}
\renewcommand{\arraystretch}{1.1}
\caption{Benchmarking PRS on different LLMs on KP (average objective and average gap (\%) to optimal). Best result among all methods of each problem size is highlighted in bold while the best result among all LLM-based results of each scale is shaded.}
\label{tab:llms_kp}

\begin{tabular}{l|cc|cc}
\noalign{\hrule height 1.2pt}
\multicolumn{5}{c}{\textbf{KP}} \\
\noalign{\hrule height 0.6pt}
\multicolumn{1}{c|}{N=} & \multicolumn{2}{c|}{\underline{N=100}} & \multicolumn{2}{c}{N=200} \\
\cline{2-5}
\multicolumn{1}{c|}{Method} & Obj. $\uparrow$ & Gap (\%) $\downarrow$ & Obj. $\uparrow$ & Gap (\%) $\downarrow$ \\
\noalign{\hrule height 0.4pt}
Optimal & 40.271 & -- & 57.448 & -- \\
POMO    & 39.676 & 1.48\% & 57.271 & 0.31\% \\
\noalign{\hrule height 0.4pt}
\multicolumn{5}{c}{\textit{Closed Source LLMs}} \\
\noalign{\hrule height 0.4pt}
GPT-4.1-nano       & 40.234 & 0.09\% & 57.405 & 0.07\% \\
GPT-5-nano         & \cellcolor{gray!20}\textbf{40.262} & \cellcolor{gray!20}\textbf{0.02\%} & \cellcolor{gray!20}\textbf{57.436} & \cellcolor{gray!20}\textbf{0.01\%} \\
Claude-3.7-sonnet  & \cellcolor{gray!20}\textbf{40.263} & \cellcolor{gray!20}\textbf{0.02\%} & 57.435 & 0.02\% \\
\noalign{\hrule height 0.4pt}
\multicolumn{5}{c}{\textit{Open Source LLMs}} \\
\noalign{\hrule height 0.4pt}
DeepSeek-v3            & 40.232 & 0.10\% & 57.398 & 0.09\% \\
Llama-3.1-70B-Instruct & 40.233 & 0.09\% & 57.400 & 0.08\% \\
\noalign{\hrule height 1.2pt}
\end{tabular}
\end{table}


\section{Cost Analysis}
\label{sec:cost_analysis}

This section reports the computational cost of different AHD frameworks. Specifically, for each framework, we measure (i) total wall-clock time, (ii) input and output token consumption, and (iii) the number of LLM queries. These metrics complement solution quality by characterizing the practical efficiency and cost profile of different frameworks under the same evaluation budget.

\begin{table}[H]
\centering
\small
\setlength{\tabcolsep}{1pt}
\renewcommand{\arraystretch}{1.05}
\caption{Cost comparison across different AHD frameworks.}
\label{tab:cost_analysis}
\begingroup
\aboverulesep=0pt
\belowrulesep=0.20pt

\begin{tabular}{lGlGcGcGcGc}
\toprule[0.5mm]
Method & Metric & OBPP-Constructive & KP-Constructive & CVRP-ACO & OP-ACO \\
\midrule[0.3mm]
\multirow{4}{*}{ReEvo}
& Time & 8h & - & 4h & 5h \\
& Input Tokens & 2.94M & - & 3.30M & 2.82M \\
& Output Tokens & 1.02M & - & 1.65M & 1.68M \\
& \#LLM Queries & 2869 & - & 1433 & 1410 \\
\midrule
\multirow{4}{*}{EoH}
& Time & 10h & 9h & 6h & 6h \\
& Input Tokens & 2.40M & 1.80M & 1.83M & 2.94M \\
& Output Tokens & 0.90M & 0.75M & 1.44M & 1.26M \\
& \#LLM Queries & 2082 & 1054 & 1045 & 1063 \\
\midrule
\multirow{4}{*}{MCTS-AHD}
& Time & 26h & 19h & 12h & 10h \\
& Input Tokens & 3.09M & 2.10M & 3.06M & 3.36M \\
& Output Tokens & 1.05M & 1.11M & 2.16M & 2.10M \\
& \#LLM Queries & 4012 & 2010 & 2004 & 2012 \\
\midrule
\multirow{4}{*}{PRS}
& Time & 10h & 5h & 6h & 5h \\
& Input Tokens & 5.00M & 3.15M & 3.09M & 3.12M \\
& Output Tokens & 2.86M & 1.87M & 1.55M & 1.60M \\
& \#LLM Queries & 2706 & 1353 & 1353 & 1353 \\
\bottomrule
\end{tabular}
\endgroup
\end{table}

\textbf{LLM queries and token consumption.}
PRS consistently uses fewer LLM queries than MCTS-AHD across all four tasks, requiring 2706 versus 4012 queries on OBPP, and 1353 versus 2010, 2004, and 2012 queries on KP, CVRP, and OP, respectively. This reduction is approximately 32-33\% across the evaluated settings. Compared with ReEvo and EoH, however, PRS does not uniformly use fewer queries: its query cost is lower than ReEvo on the tasks where both methods are reported, but can exceed that of EoH.

Token consumption follows a different pattern. PRS incurs higher token usage than the lightweight baselines in several settings, reflecting its idea-generation and idea-guided rewrite stages. Nevertheless, its token usage is lower than that of MCTS-AHD on CVRP and OP, while its query count remains substantially lower than MCTS-AHD across all tasks.

\textbf{Runtime Cost and efficiency.}
In terms of wall-clock time, PRS is comparable to lightweight evolutionary methods and significantly faster than search-heavy approaches such as MCTS-AHD across evaluated settings, as shown in Table \ref{tab:cost_analysis}. The total computational cost is primarily governed by the maximum number of evaluations $T$ and the per-heuristic evaluation time. While $T$ is consistent for all LLM-based AHD methods, the per-heuristic evaluation time of PRS remains under control, leading to a competitive total runtime among LLM-based AHD methods. This shows that PRS stands as a practical, cost-efficient AHD framework with strong exploration capability.

\textbf{Runtime factors.}
The overall runtime is primarily dominated by heuristic evaluation rather than LLM inference, and thus scales with the complexity of the heuristics rather than the branching search structure. Heuristics with expensive operations, nested loops, or inefficient data structures can substantially increase per-evaluation runtime, even under the same evaluation budget. As discussed in Appendix \ref{sec:TSP Heuristic under Guided Local Search Framework}, PRS tends to discover heuristics with favorable runtime profiles, particularly on larger problem scales, as evidenced by the Pareto analysis of solution quality and runtime in Fig. \ref{fig:tsp_gls_pareto}.

\textbf{Search granularity.}
PRS allocates its LLM budget through bounded BFS expansion. For each selected parent, one idea operator produces a target idea, which is then realized through multiple code rewrites with a fixed beam size. This organization concentrates several candidate realizations around a single idea-level proposal, rather than treating every LLM call as an independent search iteration. As shown in Table \ref{tab:cost_analysis}, this design yields substantially fewer total LLM queries than MCTS-AHD, while retaining a competitive end-to-end runtime.

\section{Sensitivity Analysis}
\label{sec:sensitivity_analysis}

We analyze the sensitivity of PRS to its main search-control hyperparameters on KP and OBPP under the step-by-step constructive framework. BFS beam size $\kappa$ and tree depth $d$ jointly control the local expansion breadth and depth; the stagnation and reheating ratio ($C/T=\Delta/T$) determines the interval for triggering stagnation-aware restart and the subsequent operator reheating horizon; the softmax temperature $T_{\mathrm{sel}}$ governs the selection pressure on the frontier; and the UCB window size $|\mathcal{H}|$ controls the amount of recent operator feedback used for adaptive operator selection. For each setting, we report the average optimality gap and token cost in Table~\ref{tab:sensitivity_analysis}. 

\begin{table}[H]
\centering
\footnotesize
\setlength{\tabcolsep}{2pt}
\renewcommand{\arraystretch}{1.1}
\caption{Sensitivity analysis of PRS on KP and OBPP under the step-by-step constructive framework. Lower gap is better.}
\label{tab:sensitivity_analysis}
\begin{tabular}{l c c c c c}
\toprule
Modified Parameter & Value
& \multicolumn{2}{c}{KP-Constructive}
& \multicolumn{2}{c}{OBPP-Constructive} \\
\cmidrule(lr){3-4} \cmidrule(lr){5-6}
& & Gap & \# Token (M)
& Gap & \# Token (M) \\
\midrule
--
& \begin{tabular}[c]{@{}c@{}}
$(\kappa=d=3),\ \Delta/T=0.1$ \\
$T_{\mathrm{sel}}=1.0,\ |\mathcal{H}|=50$
\end{tabular}
& 0.024\% & 4.25 & 0.802\% & 6.07 \\
\midrule
beam/depth
& $(\kappa=d=4)$
& 0.026\% & 3.89 & 0.835\% & 5.72 \\

beam/depth
& $(\kappa=d=2)$
& 0.031\% & 4.98 & 0.913\% & 6.35 \\
\midrule
Stagnation ratio $\Delta/T$
& 0.05
& 0.027\% & 4.13 & 0.894\% & 6.18 \\

Stagnation ratio $\Delta/T$
& 0.20
& 0.023\% & 4.40 & 0.836\% & 6.36 \\
\midrule
Softmax temperature $T_{\mathrm{sel}}$
& 0.5
& 0.021\% & 3.91 & 0.884\% & 5.86 \\

Softmax temperature $T_{\mathrm{sel}}$
& 2.0
& 0.027\% & 4.06 & 0.821\% & 5.94 \\
\midrule
UCB window $|\mathcal{H}|$
& 100
& 0.025\% & 4.29 & 0.793\% & 6.02 \\
\bottomrule
\end{tabular}
\end{table}

\textbf{Beam/depth.}
The beam/depth setting affects the local expansion breadth and depth of the search. Following the BFS convention in ToT~\citep{yao2023tree}, we adopt a bounded tree search and set $\kappa=d=3$ as the default for the BFS-based heuristic search.
In Table~\ref{tab:sensitivity_analysis}, increasing $(\kappa,d)$ from $3$ to $4$ yields comparable performance with reduced token cost, suggesting that the default setting already provides sufficient search breadth and that further expansion does not necessarily improve exploration quality within the same evaluation budget $T$. By contrast, reducing them to $2$ leads to a moderate degradation in solution quality together with a noticeable increase in token cost. This indicates that overly small search breadth or depth may hurt search efficiency, while the default setting $(\kappa=d=3)$ provides a good balance between exploration and cost.

\textbf{Stagnation ratio.}
The stagnation ratio controls both the evaluation interval for triggering stagnation-aware restart and the subsequent operator reheating horizon. Changing $\Delta/T$ from $0.1$ to $0.05$ or $0.2$ yields only minor changes in both performance and computational cost. This suggests that the adaptive search-control mechanism is robust across a reasonable range of stagnation settings.

\textbf{Softmax temperature.}
Varying the selection temperature leads to only limited performance variation across the two tasks. Although different temperatures can achieve slightly better results on individual settings, both $T_{\mathrm{sel}}=0.5$ and $T_{\mathrm{sel}}=2.0$ remain competitive with the default $T_{\mathrm{sel}}=1.0$. Meanwhile, token usage remains relatively stable across these settings, showing that the stochastic selection mechanism does not introduce large cost fluctuations. We therefore retain $T_{\mathrm{sel}}=1.0$ as a moderate default to balance exploration and exploitation within each frontier in BFS.

\textbf{UCB window size.}
Increasing the UCB reward window from $|\mathcal{H}|=50$ to $100$ results in only minor changes in both performance and token cost on KP and OBPP. This indicates that adaptive operator selection is not highly sensitive to the exact amount of recent operator feedback retained in the window. Overall, the sensitivity analysis confirms that PRS is robust to moderate variations in its main search-control hyperparameters.

\newpage

\section{Example Response for Multi-directional Perturbations}
\label{sec:example of evolution}



This section presents representative examples showing how parent heuristics are perturbed and rewritten through complementary directional transformations in PRS, as shown in Fig.~\ref{fig:expansion_example} and \ref{fig:compression_example}. For OBPP, lower scores are preferred; for OP, higher scores are better. 

\lstdefinestyle{prsstyle}{
    language=Python,
    basicstyle=\ttfamily\fontsize{6}{7}\selectfont,
    breaklines=true,
    breakatwhitespace=false,
    columns=fullflexible,
    keepspaces=true,
    showstringspaces=false,
    aboveskip=0pt,
    belowskip=0pt,
    xleftmargin=0pt,
    xrightmargin=0pt
}

\begin{figure}[H]
\centering

\begin{tcolorbox}[
    enhanced,
    width=0.88\linewidth,
    colback=black!1,
    colframe=blue!20,
    boxrule=0.6pt,
    arc=1.5mm,
    left=1.5mm,
    right=1.5mm,
    top=1mm,
    bottom=1mm,
    boxsep=0.8mm,
    title={\small\textbf{Parent Heuristic of OP-ACO} (Score: $g=13.560$)},
    fonttitle=\bfseries,
]
\begin{lstlisting}[style=prsstyle]
def heuristics(prize, distance, maxlen):
    import numpy as np
    n = len(prize)
    epsilon = 1e-8
    alpha = 1.0
    beta = 2.0
    heuristic_matrix = np.zeros((n, n))
    
    for i in range(n):
        for j in range(n):
            if i == j or distance[i][j] == 0:
                heuristic_matrix[i][j] = 0.0
            else:
                base_ratio = prize[j] / (distance[i][j] + epsilon)
                scale_factor = 1 + alpha * (
                    maxlen / (distance[i][j] + epsilon))**beta
                heuristic_matrix[i][j] = base_ratio * scale_factor
    
    row_sums = np.sum(heuristic_matrix, axis=1, keepdims=True)
    row_sums[row_sums == 0] = 1.0
    heuristic_matrix = heuristic_matrix / row_sums
    return heuristic_matrix
\end{lstlisting}
\end{tcolorbox}


\vspace{-1.2mm}
\noindent\hfill
\begin{tikzpicture}[baseline]
\draw[->, line width=1.0pt, color=green!40!black] (0,0) -- (0,-0.42)
node[midway, right=4pt, font=\footnotesize\itshape, color=green!35!black]
{Expansion: new reward-aware structures injected};
\end{tikzpicture}
\hfill\mbox{}
\vspace{-0.5mm}

\begin{tcolorbox}[
    enhanced,
    width=0.88\linewidth,
    colback=black!1,
    colframe=offspringblue,
    boxrule=0.6pt,
    arc=1.5mm,
    left=1.5mm,
    right=1.5mm,
    top=1mm,
    bottom=1mm,
    boxsep=0.8mm,
    title={\small\textbf{Offspring Heuristic of OP-ACO} (Score: $g=14.194$)},
    fonttitle=\bfseries,
]
\begin{lstlisting}[style=prsstyle]
def heuristics(prize, distance, maxlen):
    import numpy as np
    n, epsilon = len(prize), 1e-8
    gamma = delta = 1.0
    median_dist = np.median(distance)

    local_density = np.zeros(n)
    for j in range(n):
        neighbors = [k for k in range(n) if distance[j][k] <= median_dist and j != k]
        local_density[j] = np.mean(prize[neighbors]) if neighbors else prize[j]

    max_density = max(np.max(local_density), 1.0)
    local_density /= max_density

    heuristic_matrix = np.zeros((n, n))
    top_prize_sum = np.sum(prize[np.argsort(-prize)[:3]])
    total_prize = np.sum(prize)
    LAF = 1 + delta * (top_prize_sum / total_prize)

    for i in range(n):
        for j in range(n):
            if i == j or distance[i][j] == 0:
                continue
            ratio = prize[j] / (distance[i][j] + epsilon)
            scale_factor = 1 + (maxlen / (distance[i][j] + epsilon))**2
            density_weight = 1 + gamma * local_density[j]
            heuristic_matrix[i][j] = ratio * scale_factor * density_weight * LAF

    row_sums = np.sum(heuristic_matrix, axis=1, keepdims=True)
    row_sums[row_sums == 0] = 1.0
    heuristic_matrix /= row_sums
    heuristic_matrix[heuristic_matrix < 0] = 0.0
    heuristic_matrix[~np.isfinite(heuristic_matrix)] = 0.0
    return heuristic_matrix
\end{lstlisting}
\end{tcolorbox}

\caption{Example of an expansion operation on heuristics of OP under the ACO framework.
The parent heuristic (top) primarily scores transitions using prize-to-distance preference and a route-length scaling factor.
After expansion and LLM-based rewriting, the offspring (bottom) retains this backbone while introducing additional reward-aware structures, including local prize density and global high prize information.
These additions enrich the heuristic's decision logic and improve the validation score from $g=13.560$ to $g=14.194$ ($+4.68\%$).}
\label{fig:expansion_example}
\vspace{-2mm}
\end{figure}

\newpage

\begin{figure}[H]
\centering

\begin{tcolorbox}[
    enhanced,
    width=0.88\linewidth,
    colback=black!1,
    colframe=blue!20,
    boxrule=0.6pt,
    arc=1.5mm,
    left=1.5mm,
    right=1.5mm,
    top=1mm,
    bottom=1mm,
    boxsep=0.8mm,
    title={\small\textbf{Parent Heuristic of OBPP-Constructive} (Score: $g=2.170$)},
    fonttitle=\bfseries
]
\begin{lstlisting}[style=prsstyle]
def priority(item, bins_remain_cap):
    import numpy as np
    cap = np.array(bins_remain_cap, copy=True)
    epsilon = 1e-8
    max_cap = np.max(cap)
    variance = np.var(cap) / (np.sum(cap) + epsilon)
    avg_residual = np.mean(cap / (max_cap + epsilon))

    if avg_residual > 0.75:
        threshold_factor = 1.5
    elif avg_residual < 0.3:
        threshold_factor = 3.0
    else:
        threshold_factor = 2.0

    diff = cap - item
    fit_threshold = threshold_factor * item
    near_fit_mask = diff <= fit_threshold
    diff_range = (np.max(diff) - np.min(diff)) / (item + epsilon)
    disparity_factor = 1 / (diff_range + epsilon)

    neighbor_penalty = np.ones_like(cap)
    for i in range(len(cap) - 1):
        gap = np.abs(cap[i] - cap[i + 1])
        if gap > 0.1 * item:
            penalty = 1 - gap / (max_cap + epsilon)
            neighbor_penalty[i] *= penalty
            neighbor_penalty[i + 1] *= penalty

    bonus = np.zeros_like(cap)
    if np.sum(diff <= 1e-3) < 2:
        bonus[diff <= 1.2 * item] = 5.0
    else:
        bonus[near_fit_mask] = 1.0

    scores = (1 / (diff + epsilon) + bonus) * disparity_factor
    scores *= (1 - variance) * neighbor_penalty
    return scores
\end{lstlisting}
\end{tcolorbox}

\vspace{-1.2mm}
\noindent\hfill
\begin{tikzpicture}[baseline]
\draw[->, line width=1.0pt, color=green!40!black] (0,0) -- (0,-0.42)
node[midway, right=4pt, font=\footnotesize\itshape, color=green!35!black]
{Compression: simplify adaptive fit and global weighting};
\end{tikzpicture}
\hfill\mbox{}
\vspace{-1.2mm}

\begin{tcolorbox}[
    enhanced,
    width=0.88\linewidth,
    colback=black!1,
    colframe=offspringblue,
    boxrule=0.6pt,
    arc=1.5mm,
    left=1.5mm,
    right=1.5mm,
    top=1mm,
    bottom=1mm,
    boxsep=0.8mm,
    title={\small\textbf{Offspring Heuristic of OBPP-Constructive} (Score: $g=1.846$)},
    fonttitle=\bfseries,
]
\begin{lstlisting}[style=prsstyle]
def priority(item, bins_remain_cap):
    import numpy as np
    cap = np.array(bins_remain_cap, copy=True)
    epsilon = 1e-8
    max_cap = np.max(cap)
    avg_residual = np.mean(cap / (max_cap + epsilon))

    if avg_residual > 0.75:
        tight_weight = 2.0
    elif avg_residual < 0.3:
        tight_weight = 0.5
    else:
        tight_weight = 1.0

    diff = cap - item
    near_fit_mask = diff <= 1.5 * item
    bonus = np.zeros_like(cap)
    bonus[diff <= (1.2 * item if np.sum(near_fit_mask) < 2 else 1.5 * item)] = (
        3.0 if np.sum(near_fit_mask) < 2 else 1.0
    )

    disparity_factor = 1 / (
        (np.max(diff) - np.min(diff)) / (item + epsilon) + epsilon
    )
    neighbor_penalty = np.ones_like(cap)
    for i in range(len(cap) - 1):
        gap = np.abs(cap[i] - cap[i + 1])
        if gap > 0.1 * item:
            penalty = 1 - gap / (max_cap + epsilon)
            neighbor_penalty[i:i + 2] *= penalty

    scores = (1 / (diff + epsilon) + bonus) * disparity_factor
    scores *= tight_weight * neighbor_penalty
    return scores
\end{lstlisting}
\end{tcolorbox}

\caption{Example of a compression operation on heuristics of OBPP under the step-by-step constructive framework.
The parent heuristic (top) combines adaptive fit thresholds, variance-based global weighting, disparity control, and neighborhood smoothness.
After compression and LLM-based rewriting, the offspring (bottom) retains the core fit, disparity, and smoothness mechanisms while simplifying the adaptive threshold and global weighting logic.
The resulting heuristic improves the validation score from $g=2.170$ to $g=1.846$.}
\label{fig:compression_example}
\vspace{-2mm}
\end{figure}

\newpage
\section{Limitations}
\label{sec:limitations}

A current limitation of PRS is that its adaptive exploration controller still relies on predefined scheduling rules and stagnation thresholds. Although the UCB-based operator selection adapts to recent search outcomes, the overall exploration schedule and restart behavior are manually specified, which may limit the flexibility of search control across different tasks and search stages.

Another limitation is the cost of heuristic evaluation. PRS currently realizes each perturbed heuristic idea into executable code before evaluating its performance, even though not every proposed idea is necessarily worth implementation and expensive downstream evaluation. This issue can become particularly important when heuristic evaluation involves large problem instances or computationally costly optimization environments.

In future work, we aim to address these limitations by (1) developing more automatic and LLM-driven control mechanisms, where operator selection, exploration scheduling, and stagnation response are determined autonomously by the LLM agent rather than through predefined rules, and (2) introducing surrogate models, such as learned world models, to estimate the potential utility of perturbed heuristic ideas before realization. Such models could prioritize promising ideas and filter less useful ones, reducing unnecessary implementation and evaluation costs and improving the overall efficiency of heuristic search.

\section{Baselines \& Licenses}
\label{sec:licenses}

\textbf{Optimal Solutions}. For the results of TSP in Table \ref{tab:tsp_step_by_step} and Table \ref{tab:tsp_gls_ahd}, we obtain reference solutions using the Lin–Kernighan heuristic \citep{lin1973effective} implemented in the \texttt{elkai} solver, which is a widely used Python implementation of LK-based TSP solvers. 
For problem sizes up to $N=315$, the solver returns proven optimal solutions, while for larger instances it provides high-quality approximate solutions. 
We report optimality gaps with respect to these reference solutions, following common practice in prior TSP literature. For KP instances,
we use Google OR-Tools for the optimal value.

\textbf{Other Baselines Implemented.} We implement POMO~\citep{kwon2020pomo} and DeepACO~\citep{ye2023deepaco} as representative baselines. 
To ensure a fair comparison, data augmentation in POMO is disabled in all experiments. 
For KGLS, we adopt the ACO implementation provided in DeepACO and set the heuristic matrix to the distance matrix. The licenses and URLs of baselines, models, and software are summarized in Table \ref{tab:external_resources}.

\begin{table}[H]
\centering
\small
\setlength{\tabcolsep}{2pt}
\renewcommand{\arraystretch}{1.15}
\caption{External codebases and datasets used in this work.}
\label{tab:external_resources}

\begin{tabular}{l l p{0.6\columnwidth}}
\toprule
\textbf{Resource} & \textbf{License / Availability} & \textbf{URL} \\
\midrule
LKH3 & Academic research use &
\url{http://webhotel4.ruc.dk/~keld/research/LKH-3/} \\

OR-Tools & MIT License &
\url{https://developers.google.com/optimization/pack/knapsack} \\

POMO & Available online &
\url{https://github.com/yd-kwon/POMO/tree/master} \\

DeepACO & MIT License &
\url{https://github.com/henry-yeh/DeepACO} \\
\midrule
FunSearch & Apache License &
\url{https://github.com/google-deepmind/funsearch} \\

EoH & MIT License &
\url{https://github.com/FeiLiu36/EoH/tree/main} \\

ReEvo & MIT License &
\url{https://github.com/ai4co/reevo} \\

HSEvo & Available online &
\url{https://github.com/datphamvn/HSEvo} \\

MCTS-AHD & MIT License &
\url{https://github.com/zz1358m/MCTS-AHD-master} \\

\midrule
TSPLIB & Non-commercial use &
\url{http://comopt.ifi.uni-heidelberg.de/software/TSPLIB95} \\
\bottomrule
\end{tabular}
\end{table}

\end{document}